\documentclass[twocolumn,10pt,a4paper]{article}
\usepackage{naturestyle}

\newif\ifappendixatend
\appendixatendtrue

\hypersetup{%
  pdftitle={RadPRISM: Schema-stratified radiology-report supervision for concept-disentangled image representations and visual grounding},
  pdfauthor={Fabian Drexel et al.}}

\begin{document}

\twocolumn[{%
  \begin{center}
    {\sffamily\bfseries\fontsize{17}{21}\selectfont
      RadPRISM: Schema-stratified radiology-report supervision for
      concept-disentangled image representations and visual grounding\par}
    \vspace{1.0em}
    {\normalsize
      Fabian Drexel\textsuperscript{1,2,*},
      Marlene Fritzsche\textsuperscript{2},
      Era Stambollxhiu\textsuperscript{2},
      Miriam Kumpf\textsuperscript{2},
      Lena Schmitzer\textsuperscript{2},
      Lea Schumann\textsuperscript{2},
      Jannik Kahmann\textsuperscript{2},
      Friedrich Puttkammer\textsuperscript{2,3},
      Johannes Moll\textsuperscript{1,2},
      Jannik L\"ubberstedt\textsuperscript{2},
      Zeineb Ben Chaaben\textsuperscript{2},
      Anirudh Narayanan\textsuperscript{2,3},
      Cosmin I. Bercea\textsuperscript{2},
      Sebastian Ziegelmayer\textsuperscript{2},
      Marcus R. Makowski\textsuperscript{2},
      Daniel Rueckert\textsuperscript{1,4,5},
      Lisa C. Adams\textsuperscript{2},
      Keno K. Bressem\textsuperscript{2,6,7,8}\par}
    \vspace{0.7em}
    {\footnotesize\itshape
      \textsuperscript{1}Chair for AI in Healthcare and Medicine, Technical University of Munich (TUM) and TUM University Hospital, Munich, Germany;
      \textsuperscript{2}Technical University of Munich, School of Medicine and Health, TUM University Hospital, Klinikum rechts der Isar, Department of Diagnostic and Interventional Radiology, Munich, Germany;
      \textsuperscript{3}Department of Radiology, Charit\'e -- Universit\"atsmedizin Berlin, corporate member of Freie Universit\"at Berlin and Humboldt Universit\"at zu Berlin, Berlin, Germany;
      \textsuperscript{4}Department of Computing, Imperial College London, London, UK;
      \textsuperscript{5}Munich Center for Machine Learning (MCML), Munich, Germany;
      \textsuperscript{6}Institute for Artificial Intelligence in Medicine (IKIM), University Hospital Essen (A\"oR), Essen, Germany;
      \textsuperscript{7}Institute of Interventional and Diagnostic Radiology and Neuroradiology, University Hospital Essen, Essen, Germany;
      \textsuperscript{8}National Center for Tumor Diseases West, Essen, Germany.\par}
    \vspace{0.4em}
    {\footnotesize\textsuperscript{*}Corresponding author:
      \href{mailto:fabian.drexel@tum.de}{\texttt{fabian.drexel@tum.de}}\par}
  \end{center}
  \vspace{0.5em}
  {\color{black!55}\hrule height 0.6pt}
  \vspace{0.8em}
  \noindent\begin{minipage}{\textwidth}
    \small
    \textbf{\sffamily Abstract}\quad
    Vision-language pretraining learns rich medical image representations from radiology reports, but previous model variants commonly operate within a single shared embedding space, so concept-level structure and interpretability must be recovered post hoc, limiting model transparency and, hence, clinical utility. We introduce RadPRISM, which makes a clinician-defined radiology schema a designated stratification axis: an on-premise large language model extracts per-concept text spans from free-text reports, and each clinical concept is aligned in its own dedicated visual subspace, turning concept stratification into direct, top-level alignment supervision. Instantiated on chest radiographs with a 19-concept schema over $203{,}602$ examinations from an internal multi-year archive, RadPRISM improved internal dataset zero-shot classification from $0.717$ (95\% CI, $0.710-0.723$) to $0.868$ (95\% CI, $0.863-0.872$) macro AUROC over a matched global-alignment baseline, performed on par with the purpose-built CARZero reference in external zero-shot classification while substantially outperforming it (up to 4.3-fold) in pointing-game visual grounding. In addition, a radiologist reader study demonstrated concept-stratified retrieval ability ($0.78$ macro retrieval correctness rate within rank 3), surfacing disentangled descriptive findings that report-level retrieval and fixed-label vocabularies cannot express. RadPRISM yields discriminative, spatially faithful, natively concept-stratified representations shaped by and transparently inspectable by clinicians.
  \end{minipage}
  \vspace{0.6em}
  {\color{black!55}\hrule height 0.6pt}
  \vspace{1.6em}
}]

\setcounter{secnumdepth}{0}

\section{Introduction}
\begin{figure*}[t!]
  \centering
  \includegraphics[width=\linewidth]{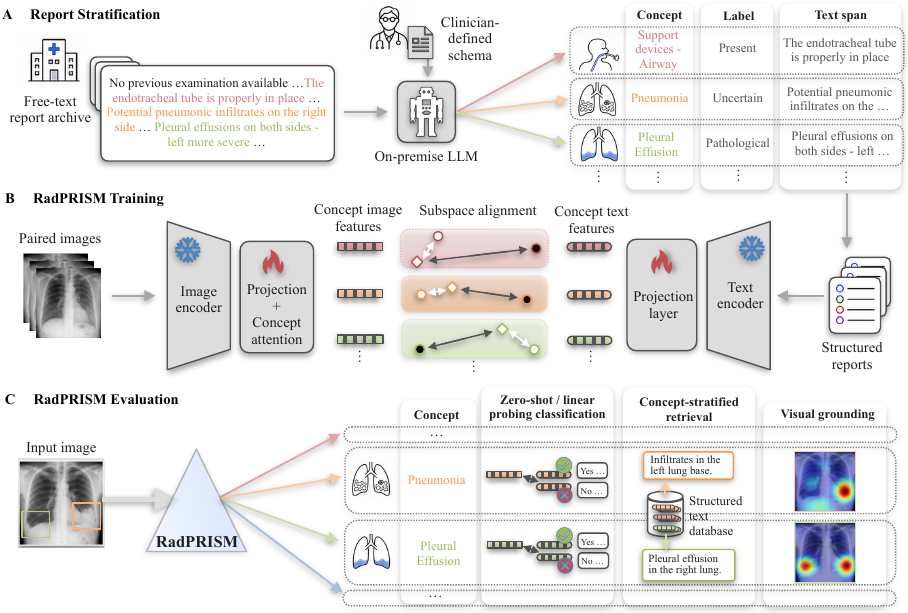}
  \caption{Overview of the schema report stratification (\textbf{A}), the RadPRISM training (\textbf{B}) and evaluation (\textbf{C})  approach, spanning LLM-based report structuring, concept-specific vision-language alignment, and downstream stratified evaluation.}
  \label{fig:structuring-vlm-eval-overview}
\end{figure*}

Radiological findings are routinely communicated through free-text reports that describe not only the presence of an abnormality but also descriptive attributes such as its location, extent, and degree of diagnostic certainty. Medical artificial intelligence for image interpretation, by contrast, has been developed predominantly within a closed-vocabulary supervision paradigm, in which a fixed set of disease labels is curated and used to train classifiers against expert-annotated examples.\cite{tiu2022chexzero} This paradigm has produced strong narrow models but constrains the resulting representations to the granularity of the chosen label set, collapsing graded clinical reasoning, such as the distinction between a definite finding and one that cannot be excluded, into a binary presence-or-absence target and leaving any descriptive content beyond the vocabulary structurally absent from the learned features. The free-text radiology report, which, by design, carries this graded, descriptive content, is generated as a routine product of clinical care and accumulates at scale in hospital archives, yet remains largely unexploited as a customized training signal within many institutions.

Contrastive vision–language (VL) pretraining provides a route to exploit this signal. Zhang et al. first demonstrated that aligning a single global image embedding with the corresponding full-report embedding yields useful chest radiograph representations without disease-level annotation,\cite{zhang2022contrastive} and Tiu et al. showed that a sufficiently scaled instance of this paradigm enables expert-level zero-shot multi-label classification across an external test set, including for pathology labels never explicitly seen during training.\cite{tiu2022chexzero} Subsequent extensions to volumetric imaging using CT-report pairs followed the same alignment principle.\cite{hamamci2024ctrate}
A second line of work introduced alignment objectives that decompose the report or image into smaller components — sentences, clinical entities, or local regions — rather than treating the report as a whole.\cite{huang2021gloria,boecking2022biovil,wu2023medklip,zhang2023kad,park2026radzero} Across these approaches, however, alignment is still performed within a single shared embedding space, so per-concept structure is not enforced by the supervisory signal itself.

Two consequences arise from this geometry that existing methods do not address. Sub-categorical descriptive content within a clinical concept, such as the specific subtype of a support device or the qualifier accompanying a tentative finding, must be recovered indirectly from the surrounding text, and any concept-level structure in the resulting representation is whatever the embedding space happens to encode, not a property the supervisory signal enforces.\cite{imran2026multimodal} Beyond limiting interpretability, this geometry also constrains what the supervisory signal itself can teach the model. With one image feature vector aligned to one report text-derived feature vector, the contrastive gradient is shaped by whichever aspects of the report most strongly distinguish it from negatives, and descriptive variants or subtle sub-categorical content that occupy only a small portion of any single report contribute little to learning. Sub-categorical content and the broad spectrum of clinically relevant descriptors are accordingly under-represented in the learned features, irrespective of the size of the training corpus.

Closing this gap requires a representation in which each relevant clinical concept occupies its own visual subspace and is supervised by the descriptive text spans that carry its sub-categorical detail and qualifiers, so that this content enters the training signal directly. Prior stratified approaches have moved in this direction but rely on short class names or labels as per-concept supervision and recombine concept outputs into report-level objectives,\cite{gu2025radalign,gu2026anatomy} leaving the per-concept training signal coarse and the representation only partially stratified. We address this with RadPRISM (Radiology Per-concept Representation learning via Interpretable Stratified Multimodal alignment), a schema-driven concept-stratified VL framework that takes a clinician-defined radiology reporting schema as its supervisory axis. Free-text reports are decomposed by an open-weight large language model (LLM) into per-concept descriptive spans, and parallel concept-wise contrastive losses align a dedicated visual subspace for each concept to its corresponding text. The resulting representation is stratified by construction, outperforming a global-alignment baseline trained on the same reports with the same encoders. RadPRISM supports per-concept classification, visual grounding that partly reaches human-level performance on an external benchmark without any localization supervision during training, and concept-disentangled text retrieval of descriptive content beyond what fixed-label supervision can express. We evaluate the framework on chest radiography across an internal multi-year archive against a zero-shot and label-supervised linear-probing baseline and on the public CheXlocalize benchmark, \cite{saporta2022benchmarking} against the zero-shot cross-attention reference CARZero.\cite{lai2024carzero} We then complement these analyses with a radiologist reader study spanning all 19 schema concepts. An overview of the framework is provided in Fig.~\ref{fig:structuring-vlm-eval-overview}.

\section{Results}
\subsection*{Dataset and LLM-based report structuring}

The framework was developed on a multi-year archive of 323{,}562 chest radiographs from 203{,}602 examinations of 114{,}573 patients (55\% male, mean age $63 \pm 18$ years) acquired at \emph{Technical University of Munich (TUM) University Hospital} (TUM UH) between January 2005 and March 2024, paired with the corresponding free-text radiology reports. Approximately 56\% of examinations included paired frontal and lateral projections and 42\% a single frontal projection, with dataset splits defined at the patient level to prevent near-duplicate leakage.
A locally hosted open-weight LLM (GPT-OSS-120B~\cite{openai2025gptoss120bgptoss20bmodel}) decomposed each report into per-concept text spans and discrete labels conforming to a clinician-defined 19-concept schema (designed by two board-certified radiologists) spanning thoracic organs, pathology categories, and support devices (Fig.~\ref{fig:structuring-vlm-eval-overview} \textbf{A}). The prespecified schema-extracted spans and labels provided the per-concept supervision for the proposed RadPRISM model.

To verify that the LLM-extracted supervision faithfully reflected the underlying reports, three radiology residents corrected structured outputs across 200 reports drawn from the processed report pool.
Label fidelity tracked human consensus closely, with a median macro $F_1$ of 0.89 (range: 0.71--0.99) across all labeled concept fields (Fig.~\ref{fig: internal_testset_results_figure} \textbf{A}) and inter-rater reliability of Krippendorff's $\alpha = 0.953$ for binary fields and $\alpha = 0.863$ for multiclass fields, both within the substantial-to-near-perfect range conventionally used for inter-rater agreement \cite{krippendorff2011computing}. Cohen's $\kappa$ values followed the same trend (Supplementary Table~\ref{tab:reader-study-metrics}), and per-concept inter-rater reliability values are reported in Supplementary Table~\ref{tab:irr-per-field}. 
Text span extraction reached a median macro BERTScore $F_1$ of 0.88 (range: 0.37--1.00), indicating high semantic concordance with the corrected references for the majority of defined concepts (Fig.~\ref{fig: internal_testset_results_figure} \textbf{A}).
Residual extraction errors clustered on concept fields where the radiologists themselves disagreed, with pairwise human inter-reader BERTScore $F_1$ falling to 0.36 on \textit{Vessels: Pulmonary edema} (Supplementary Table~\ref{tab: appendix_interreader_bertscore_f1}). Human inter-reader agreement, therefore, bounds the achievable structuring quality assessment in these fields.

\subsection*{Concept-stratified representation learning}

\begin{figure*}[t!]
  \centering
  \includegraphics[width=\linewidth]{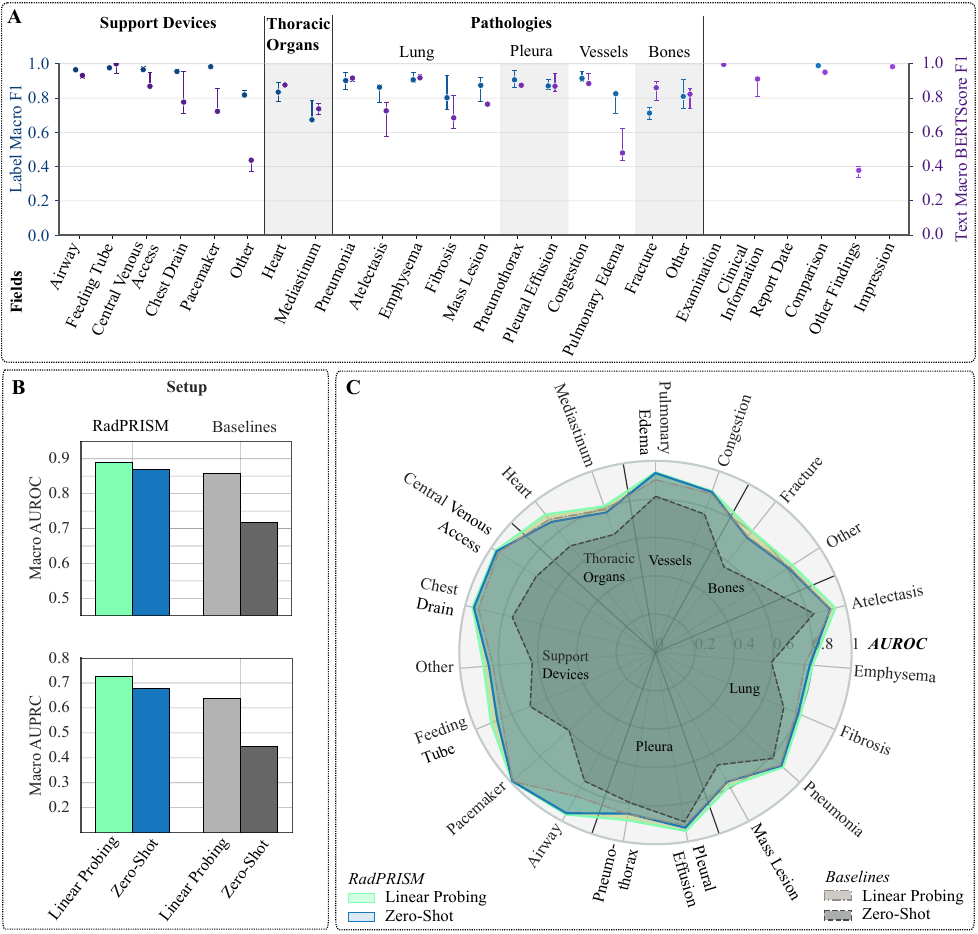}
  \caption{\textbf{A} Structuring and labeling quality reader study results. Displayed are the macro F1 scores for the different structuring template label fields and the extracted text macro BERTScore F1 value for all fields (median with min-max whiskers). The report date extraction performance score is missing, as there were no occurrences in the reader study sample pool. \textbf{B} Macro AUROC and AUPRC RadPRISM classification performance on the internal dataset test split with comparison to the baseline setups. \textbf{C} Concept-wise AUROC classification deep-dive for RadPRISM and the baseline setups. The \textit{Bones: Other} concept captures aspects like degeneration, scoliosis, etc..}
  \label{fig: internal_testset_results_figure}
\end{figure*}

With matched encoders, projection/attention adapters, and training data, schema-driven, concept-stratified supervision substantially outperformed global, unstructured report alignment on internal zero-shot classification. RadPRISM reached macro AUROC 0.868 (95\% CI, $0.863-0.872$; macro AUPRC 0.676 with 95\% CI, $0.669-0.684$), compared to 0.717 (95\% CI, $0.710-0.723$; macro AUPRC 0.451 with 95\% CI, $0.443-0.461$) for the unstructured-report baseline, which aligned a global image embedding with the corresponding report embedding as in standard contrastive alignment frameworks (Fig.~\ref{fig: internal_testset_results_figure} \textbf{B}). To estimate how much of the concept-relevant information was carried by the learned visual representation, a linear classifier was fit per concept on top of the frozen RadPRISM features. This linear-probing setup reached macro AUROC 0.890 (95\% CI, $0.886-0.894$; macro AUPRC 0.725 with 95\% CI, $0.717-0.733$), exceeding a label-supervised baseline trained directly on the frozen ViT CLS token (macro AUROC 0.859, with 95\% CI, $0.854-0.864$; macro AUPRC 0.642, with 95\% CI, $0.634-0.649$) on the same data. The baseline setups and parameters used are detailed in the Methods (Baseline and ablation setups) and Supplementary Section~\ref{app:baselines-details}. Per-seed variability across three independent runs was an order of magnitude smaller than the observed gaps for all reported setups (details in Supplementary Section~\ref{subsubsec: appendix_internal_eval_extended}). 

RadPRISM per-concept zero-shot AUROC across the 19 concepts ranged from 0.759 on \textit{Bones: Fracture} to 0.992 on \textit{Pacemaker} (Fig.~\ref{fig: internal_testset_results_figure} \textbf{C}), leading across all concepts compared to the zero-shot baseline. The highest performance clustered on support devices (\textit{Pacemaker} 0.992, \textit{Central venous access} 0.967, \textit{Chest drain} 0.955, \textit{Airway devices} 0.953), with \textit{Pulmonary edema} (0.936) the strongest pathology category. The lowest performance fell on \textit{Fracture} (0.759), \textit{Mass lesion} (0.767), \textit{Mediastinum} (0.772), \textit{Emphysema} (0.790), and \textit{Fibrosis} (0.794). Within this group, \textit{Fibrosis} and \textit{Emphysema} carried the highest training-set masking rates (95.1\% and 91.5\%; Supplementary Table~\ref{tab:text_label_split_overview_after_65pct_imputation}), while \textit{Fibrosis} (0.680), \textit{Mass lesion} (0.717), and \textit{Fracture} (0.753) ranked among the lower inter-reader BERTScore $F_1$ fields in the structuring reader study (Supplementary Table~\ref{tab: appendix_interreader_bertscore_f1}). Linear-probing gains for RadPRISM over zero-shot were negligible for already-saturated concepts ($\Delta$AUROC $\leq 0.005$ for \textit{Pacemaker}, \textit{Congestion}, and \textit{Pulmonary edema}) and substantial for \textit{Heart} ($\Delta$AUROC $=0.049$) and \textit{Fracture} ($\Delta$AUROC $=0.044$). Further per-concept difference details are reported in Supplementary Section~\ref{subsubsec: appendix_internal_eval_extended}. 

\subsection*{Generalization to external data}

\begin{figure*}[t!]
  \centering
  \includegraphics[width=\linewidth]{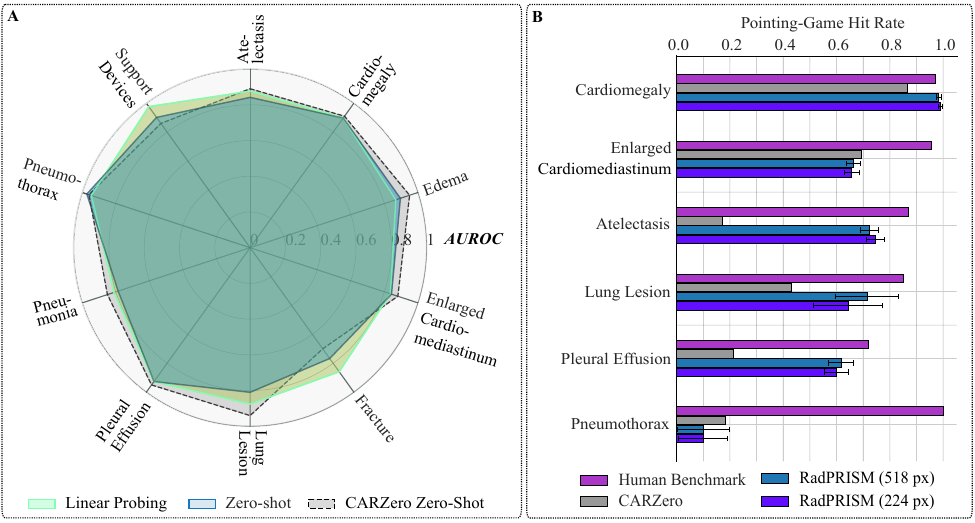}
  \caption{\textbf{A} RadPRISM concept-wise zero-shot and linear probing classification performance on the CheXlocalize test split with CARZero reference. \textbf{B} RadPRISM (224~px and 518~px image resolution variant) visual grounding performance evaluated via pointing game hit rate (whiskers represent bootstrap-estimated standard deviations) on the CheXlocalize test split, including CARZero scores and human benchmark values from Saporta et al.~\cite{saporta2022benchmarking}.}
  \label{fig: external_testset_results_figure}
\end{figure*}
 
To assess whether RadPRISM performance transfers beyond the training institution, the model was evaluated on the public CheXlocalize benchmark \cite{saporta2022benchmarking}, a localization-annotated derivative of CheXpert \cite{irvin2019chexpert} acquired at Stanford Health Care. The evaluation imposes two simultaneous shifts relative to the training distribution. The institution and the label schema differ (CheXpert's 14 observations versus the 19-concept TUM schema). The downstream evaluation was restricted to the subset of CheXpert classes for which a mapping from the structured-report concepts was available (Supplementary Section~\ref{subsec: appendix_chexpert_label_class_concept_mappings}).

Zero-shot classification on the mapped concepts still reached high per-concept AUROC (macro AUROC 0.861, with 95\% CI, $0.824-0.900$)  despite the data shifts, with the strongest performance on \textit{Pneumothorax} (0.973), \textit{Pleural Effusion} (0.926), and \textit{Support Devices} (0.902), and the lowest on \textit{Fracture} (0.767), \textit{Pneumonia} (0.786), and \textit{Lung Lesion} (0.810) (Fig.~\ref{fig: external_testset_results_figure} \textbf{A}). Linear probing further improved per-concept performance, with the largest gains observed for \textit{Fracture} ($\Delta$AUROC $=+0.092$) and \textit{Support Devices} ($+0.075$). Compared against CARZero~\cite{lai2024carzero}, a zero-shot cross-attention VL model designed specifically for this task family, RadPRISM achieved competitive zero-shot classification across the mapped concepts, with a mean per-class difference in zero-shot AUROC of only $-0.025$ across the ten mapped concepts (at a macro AUROC upper-lower 95\% CI span of $0.076$), and the largest differences (\textit{Lung Lesion}) narrowing under linear probing.

Visual grounding, evaluated via the pointing-game hit rate, revealed the cleanest separation between RadPRISM and the cross-attention reference (Fig.~\ref{fig: external_testset_results_figure} \textbf{B}). On \textit{Cardiomegaly}, RadPRISM reached a hit rate of 0.983 (95\% CI, $0.960-1.0$) at 518~px input resolution, matching the human reference of 0.972 \cite{saporta2022benchmarking}. At matched 224~px input resolution, RadPRISM outperformed CARZero by factors of 4.3 on \textit{Atelectasis} (0.746 at 95\% CI, $0.684-0.808$ versus 0.172), 2.8 on \textit{Pleural Effusion} (0.600 at 95\% CI, $0.508-0.692$ versus 0.214), and 1.5 on \textit{Lung Lesion} (0.643 at 95\% CI, $0.429-0.857$ versus 0.429). No localization supervision was used at any stage of training, and the two resolution variants produced hit-rate metrics stable within bootstrapped uncertainty estimates. Auxiliary Dice and IoU scores are reported in Supplementary Table~\ref{tab: appendix_grounding_dice_iou}.

Two concepts diverged from this pattern. On \textit{Pneumothorax}, classification AUROC (0.973) and grounding hit rate 0.100 (95\% CI, $0.000-0.300$) separated sharply, with CARZero showing the same dissociation (hit rate 0.182). On \textit{Enlarged Cardiomediastinum}, RadPRISM reached a hit rate of 0.657 (95\% CI, $0.603-0.710$), slightly below CARZero (0.693) and well below the human reference (0.957). Both deviations are examined in the Discussion below.

\subsection*{Reader study and concept-disentangled retrieval}

\begin{figure*}[t!]
  \centering
  \includegraphics[width=\linewidth]{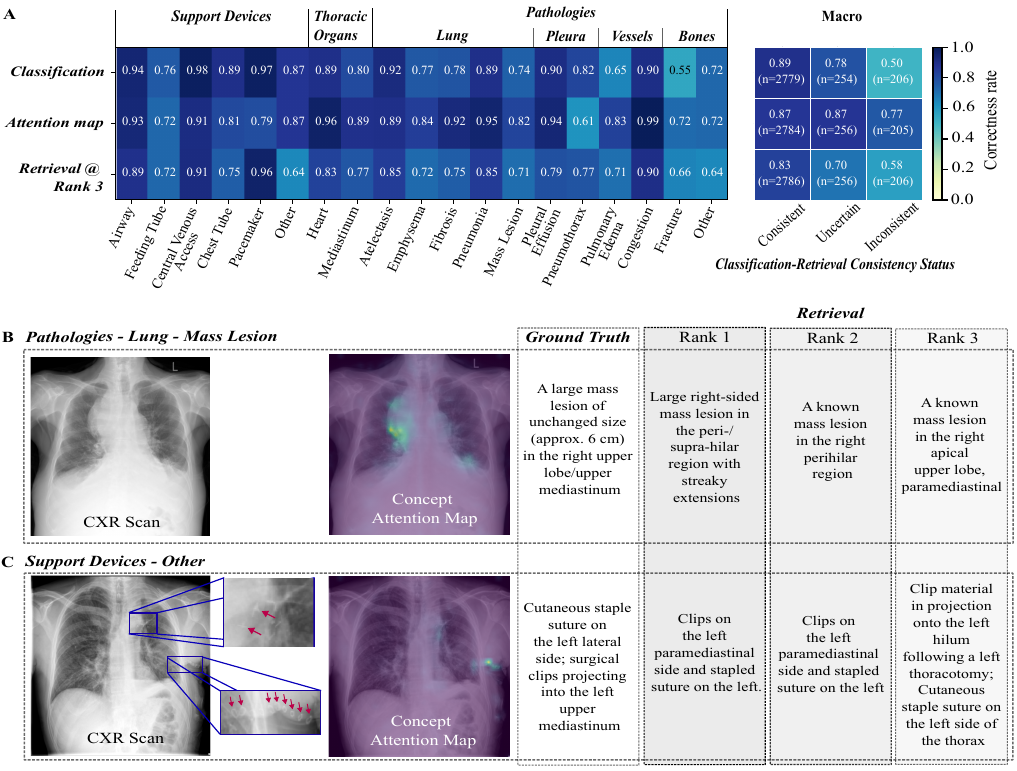}
  \caption{\textbf{A} Per-concept and macro-averaged correctness rates for binary classification, attention-map localization, and top-3 text retrieval across the 19-concept schema in the radiologist reader study, with right-panel breakdown by classification-retrieval consistency group. \textbf{B, C} Qualitative case examples showing the chest radiograph, the concept-specific attention map, the ground-truth report description, and the top three retrieved text spans from the per-concept reference database.}
  \label{fig: reader_study_retrieval_figure}
\end{figure*}

In order to extend the evaluation to all 19 schema concepts and test a capability that standard binary-label metrics cannot measure, a study involving radiologist readers was conducted on 200 chest radiographs (100 internal and 100 CheXlocalize) across three complementary axes: classification, attention-map localization, and top-3 text concept-stratified retrieval correctness. Six radiology residents assessed the cases across all three axes under the supervision of two board-certified radiologists (KKB and LCA), who had nine and ten years of cardiothoracic radiology experience, respectively. For retrieval, a per-concept reference database was constructed exclusively from training-split structured-report spans, so that the model could surface concept-specific descriptive text for any test image.

Across the 19 concepts, radiologists judged RadPRISM correct on 0.83 of cases for binary classification, 0.85 for attention-map localization, and 0.78 for top-3 text retrieval (Fig.~\ref{fig: reader_study_retrieval_figure} \textbf{A}; Correctness rates are computed over cases with existing valid reader feedback). Per-concept patterns for classification and visual grounding aligned with the quantitative evaluations established on the internal and external test splits. Classification correctness was lowest on \textit{Bones: Fracture} (0.55), \textit{Vessels: Pulmonary edema} (0.65), and \textit{Bones: Other} (0.72). Attention-map correctness was lowest on \textit{Pleura: Pneumothorax} (0.61), followed by \textit{Bones: Fracture}, \textit{Feeding tube}, and \textit{Bones: Other} (0.72 each).
Per-concept correctness rates broken down by sample pool (internal versus CheXlocalize) and extended reader feedback details are reported in Supplementary Tables \ref{tab:appendix_vlm_classification_feedback} and \ref{tab:appendix_vlm_attention_feedback}.

The retrieval axis tests a capability that fixed-label supervision is structurally unable to express. Given a test image and a target concept, RadPRISM surfaces the most similar concept-specific descriptive spans from the per-concept reference database. 
At a macro rank-3 correctness rate of 0.78, radiologists confirmed that the retrieved descriptions predominantly matched the underlying imaging finding for the queried concept, without any retrieval-specific objective, image-level annotation, or fine-tuning anywhere in the training pipeline. The retrieved spans carry imaging characteristics that lie beyond the discrete-label vocabulary, including specific support-device subtypes and pathology-specific qualifiers covering extent, laterality, and diagnostic certainty, as illustrated by the qualitative case examples in Fig.~\ref{fig: reader_study_retrieval_figure} \textbf{B, C} and additional examples in Supplementary Figs.~\ref{fig: appendix_further_attn_map_examples1} and~\ref{fig: appendix_further_attn_map_examples2}. 
The lowest per-concept retrieval correctness rates fell on \textit{Support devices: Other} and \textit{Bones: Other} (0.64 each), \textit{Bones: Fracture} (0.66), \textit{Vessels: Pulmonary edema} and \textit{Lung: Mass lesion} (0.71 each), and \textit{Lung: Emphysema} (0.72).
Per-concept retrieval correctness rates broken down by sample pool and further reader feedback details are reported in Supplementary Tables \ref{tab:appendix_vlm_retrieval_feedback} and \ref{tab:appendix_vlm_retrieval_at_k}.

Beyond per-task correctness, agreement between the binary classification and top retrieval result was inspected with respect to a potential per-concept signal of model confidence. Cases in which classification and retrieval agreed showed substantially higher radiologist-confirmed correctness on all three evaluation axes than cases in which the two disagreed or in which it was uncertain (Fig.~\ref{fig: reader_study_retrieval_figure} \textbf{A}, right panel). For binary classification, all classification-versus-retrieval consistency contrasts reached $P < 0.001$. For attention-map correctness, the consistent-versus-inconsistent contrast reached $P < 0.001$, the uncertain-versus-inconsistent contrast $P = 0.006$, the consistent-versus-uncertain contrast $P = 0.985$. For top-3 retrieval correctness, the consistent-versus-inconsistent and consistent-versus-uncertain contrasts reached $P < 0.001$, and the uncertain-versus-inconsistent contrast $P = 0.009$.
Per-concept reader consistency assessment details broken down by sample pool are reported in Supplementary Table \ref{tab:appendix_vlm_consistency}.

\section{Discussion}
In this study, RadPRISM demonstrates that decomposing free-text radiology reports along a clinician-defined schema yields a vision-language representation that is simultaneously discriminative, spatially faithful, and inspectable at the level of individual clinical findings. Trained on one of the largest single-institution CXR-report corpora assembled to date for vision-language pretraining (323{,}562 chest radiographs from 203{,}602 examinations), schema-stratified supervision outperformed a global report alignment baseline trained on the same data and encoders by 0.151 macro AUROC on internal zero-shot classification, matched the radiologist reference for visual grounding of cardiomegaly on external data without any localization supervision, and transferred zero-shot across simultaneous shifts in institution and label vocabulary. A radiologist reader study spanning all 19 concepts confirmed concept-disentangled retrieval of descriptive imaging content beyond the granularity of any fixed-label vocabulary, with per-concept agreement between classification and retrieval providing a promising uncertainty signal at the level of individual findings. By recasting the institutional free-text report archive as a directly usable supervisory resource, this work resolves a recognized limitation of global image-report alignment and outlines a route to clinical AI whose supervisory granularity scales with the descriptive language radiologists already use in practice.

The 0.151 macro AUROC advantage over the global-report alignment baseline is informative because it isolates supervisory geometry as the active ingredient. Both setups shared the same frozen image and text encoders, projection/attention adapters, the same training corpus, and the same contrastive objective, but differed in whether the report was treated as a single narrative for global alignment or as a concatenation of concept-specific spans for subspace alignment. Contrastive learning on a single image-report text vector pair is shaped by whichever aspects of the report most strongly distinguish it from negatives, so concept-specific descriptive variants, qualifiers, and sub-categorical content that occupy only a small portion of any single report contribute little to learning, a long-recognized concern of compositional reasoning and fine-grained signal capture in joint embedding spaces~\cite{imran2026multimodal, yuksekgonul2023when, parcalabescu2022valse}. Routing each of these signals into its own visual subspace prevents them from being averaged out by the bulk of the report. The alignment-pretrained representation also outperformed a directly label-supervised CLS-token baseline under linear probing, extending this picture and indicating that schema-driven alignment reorganizes the underlying foundation features such that fully supervised downstream classifiers benefit beyond zero-shot use.

Whether this advantage survives outside the institution of training is the question on which any clinical claim depends, and external evaluation on the CheXlocalize benchmark imposed two simultaneous distribution shifts to address it, spanning institution and label vocabulary~\cite{irvin2019chexpert, saporta2022benchmarking}. Zero-shot classification remained competitive with CARZero~\cite{lai2024carzero}, a state-of-the-art cross-attention VL model purpose-built for zero-shot CXR classification, and the pointing-game hit rate on \textit{Cardiomegaly} matched the radiologist reference of Saporta et al.~\cite{saporta2022benchmarking}. At matched 224\,px input resolution, grounding hit rates exceeded CARZero by factors of 4.3 on \textit{Atelectasis}, 2.8 on \textit{Pleural Effusion}, and 1.5 on \textit{Lung Lesion}, with no localization supervision used at any training stage. Concept boundaries defined in a German clinician schema, therefore, mapped onto findings annotated against a different vocabulary, and the spatial anchors learned through concept-stratified contrastive supervision generalized to an external benchmark acquired at a different institution. The schema does not bind the representation to its institution of origin, consistent with recent observations that institutionally adaptive medical foundation models benefit from supervision grounded in local clinical text rather than transferred through fixed-label vocabularies~\cite{moor2023generalist, christensen2024vision}.

Beyond aggregate metrics, the most distinctive evidence that schema-stratified supervision captures content that the discrete-label vocabulary cannot reach came from the radiologist-validated retrieval task. From a per-concept reference database constructed exclusively from training-split report spans, descriptive text matching the queried imaging finding was retrieved without any retrieval-specific objective, image-level annotation, or fine-tuning, and the retrieved spans carried information that fixed-label supervision is structurally unable to express, including specific support-device subtypes, pathology qualifiers covering extent and laterality, and degrees of diagnostic certainty. Prior stratified VL alignment work in radiology has either supervised concepts with short class descriptions rather than radiologist-authored spans directly, or recombined per-concept outputs into report-level retrieval objectives~\cite{gu2025radalign}, and neither configuration can surface sub-categorical descriptive content per concept, because the relevant signal is either absent from the supervisory text or compressed during recombination. A second consequence of report-level treatment is that models learn to prioritize narrative coherence over strict visual fidelity, a bias that LLaVA-TA~\cite{cheng2026rethinking} addresses by fine-tuning under explicit output constraints in combination with an external segmentation model, and that has also been documented for generative radiology report drafting, where hallucinated narrative is anchored more in language priors than in image evidence~\cite{miura2021improving, yu2023evaluating}. Concept-stratified alignment supervision breaks this coupling at the source, as each concept's text is presented as an independent per-concept span and representation within subspaces rather than as part of a narrative that the representation is rewarded for reproducing.

The same per-concept separation enables finding-level uncertainty signals that case-level calibration cannot provide. When the binary classification and top retrieval outputs of RadPRISM agreed on a given concept, radiologist-confirmed correctness on that concept was substantially higher across all three reader-study axes than when the two outputs disagreed, with the consistency contrasts for classification correctness reaching $P < 0.001$ and with the same direction observed on the attention-map and retrieval axes, though not all pairwise contrasts reached significance. The signal is available at inference time and operates per finding rather than per case. In workflows that triage outputs finding by finding, where the model's certainty on cardiomegaly must be available independently of its certainty on rib fracture, per-concept disagreement addresses a granularity limitation that case-level uncertainty estimates in medical AI structurally cannot~\cite{abdar2021review, kompa2021second}.

Certain concepts deviated from the discussed patterns across distinct evaluation scenarios, with failure modes traceable to various limitations. On \textit{Pneumothorax}, classification AUROC reached 0.973 while the pointing-game hit rate fell to 0.100, and the same dissociation appeared for CARZero (hit rate 0.182). The thin pleural line that defines pneumothorax approaches the spatial resolution limit at the 518\,px training input as well as at the 224\,px CARZero input, so both models likely recognized pneumothorax from contextual priors such as chest-tube presence and lung-field asymmetry rather than from the pleural line itself, consistent with the recognition-grounding gap documented for low-resolution contrastive VL alignment~\cite{luo2025xbench}. On \textit{Enlarged Cardiomediastinum}, the non-cardiac mediastinum was correctly localized at the anatomical resolution of the training schema, which defines mediastinum and heart as separate concepts, and was penalized by a benchmark annotation that combined the two. 
The first failure traces to input resolution rather than to the supervisory design, where a similar constraint is expected for \textit{Bones: Fracture}, which was the weakest concept on internal zero-shot classification, its defining fine linear discontinuities lying at the same scale and remaining difficult for human readers even at full diagnostic resolution. Spatial scale relative to input resolution, therefore, bounds per-concept performance independently of how the supervision is structured. 
The second illustrates that schema choices made at training time bound what cross-dataset evaluation can fairly measure. 
In addition, a separate source of variability lies upstream. Concepts with lower LLM structuring quality or higher inter-rater disagreement in the structuring reader study consistently ranked among the lowest across classification, grounding, and retrieval, with pairwise human inter-reader BERTScore $F_1$ on the most variable of these fields falling as low as 0.36. A substantial portion of the apparent extraction limitation, therefore, reflects interpretive ambiguity in the underlying clinical text rather than an LLM-specific failure, and the ceiling can be raised only by stricter schema definitions or by surfacing concept-level uncertainty as an explicit output of the system. This pattern is consistent with broader observations that the quality of LLM-based clinical text structuring tracks the structural clarity of the source documentation and the schema applied to it~\cite{woznicki2024structuring}.   

Two further factors shape the same spread. Concepts whose radiographic expression is diffuse or heterogeneous rather than focal and high-contrast clustered at the lower end of internal zero-shot performance, among them \textit{Mediastinum}, whose structuring agreement was comparatively high (inter-reader BERTScore $F_1$ 0.856) and whose difficulty therefore lies in the imaging appearance rather than in the supervisory text ambiguity or structuring quality. Support devices, with their high-contrast geometric signatures and formulaic report descriptions, clustered at the opposite end. Supervisory sparsity compounds this: \textit{Fibrosis} and \textit{Emphysema} carried training-set masking rates above 91\%, so for part of the low-performing group, the limiting factor was how rarely the concept was described at all across the CXR reports rather than how well it was described. Neither factor is specific to concept-stratified alignment, but together with structuring quality and spatial scale, they bound the per-concept ceiling that any schema-driven representation can reach.

Three properties of the resulting representation map onto requirements for clinical deployment that have proven difficult to satisfy in contrastive learning or label-supervised pipelines, and they follow directly from the supervisory geometry. The first is inspectability. Per-concept visual subspaces, attention maps, and retrieved descriptive text are aligned at the level of individual clinical findings rather than the case as a whole, so model behavior on one concept can be audited without entangling it with behavior on others, addressing a recognized obstacle to the regulatory and clinical acceptance of opaque imaging classifiers~\cite{ghassemi2021false, reyes2020interpretability}. The second is institutional adaptability. The framework treats the schema and the foundation encoders as interchangeable components, and a hospital's existing free-text report archive provides the per-concept supervisory text without requiring purpose-built annotation, which has been the dominant rate-limiting step in deploying clinical AI~\cite{willemink2020preparing, varoquaux2022machine}. The third is end-to-end on-premise feasibility. The LLM-based structuring step runs on a locally hosted open-weight model, and VL pretraining requires only the institution's own images and reports, with no patient-level data leaving the institutional perimeter, an increasingly relevant constraint under European medical-device and data-protection frameworks~\cite{muehlematter2021approval}. Together, these properties outline a deployment pattern in which institutionally grounded clinical AI is built from the institution's own routine clinical text, audited at the concept level by the clinicians who produce that text, and updated as the schema evolves. The 19-concept chest radiography instantiation presented here is a proof of principle for that pattern rather than its final form.

Several aspects of this study limit the conclusions that can be drawn. Training was conducted at a single institution, and although external classification, grounding, and retrieval evaluation on CheXlocalize provide indirect evidence of generalizability, prospective multi-site training and deployment data are required before any operational claim can be made. The downstream evaluation was restricted to chest radiography, and whether schema-stratified alignment scales to volumetric modalities, in which reports are longer, and findings are organized along three spatial axes, is not established here, although the principle has been pursued in initial form for CT~\cite{hamamci2024ctrate, lin2024ctglip}. Alignment training relied on weak labels and text spans extracted by an LLM rather than on radiologist-corrected supervision, and the reader-study analyses indicate that LLM structuring fidelity acts as a partial ceiling on per-concept downstream performance. The spatial recognition-grounding gap observed for \textit{Pneumothorax} persists despite schema-level supervision and is unlikely to be resolved without higher input resolution or explicit spatial regularization. The reader study was conducted at the case level by six radiology residents under the direct supervision of two board-certified radiologists with nine and ten years of cardiothoracic experience, rather than by multiple independent attendings, and the reported correctness rates should be interpreted in light of this reader-pool composition. No prospective evaluation of clinical workflow integration or patient-relevant outcomes was performed, and the deployment claims made above remain hypotheses to be tested in dedicated follow-up studies.

This work establishes that institutional free-text radiology report archives, decomposed at scale by an on-premise open-weight LLM along a clinician-defined schema, can supervise vision-language representations that are simultaneously discriminative, spatially faithful, and inspectable at the level of individual clinical concepts. The 19-concept chest radiography instantiation is a proof of principle. Its modular treatment of schema and foundation encoders as interchangeable components extends naturally to other reporting-rich modalities, with volumetric imaging as the most immediate target. The concept-disentangled retrieval signal, paired with classification-retrieval consistency as a potential per-concept uncertainty channel, offers a direct substrate for human-in-the-loop reporting workflows in which expert descriptions surface as editable proposals during drafting. With current open-weight LLMs and pretrained foundation components, the building blocks for clinically inspectable and institutionally adaptable medical AI are now in place. What remains is to demonstrate, prospectively and across modalities, that representations supervised by the descriptive language clinicians already use can carry this capability into routine care.

\section{Methods}
\subsection{Datasets}

\textbf{Internal dataset.} Reports were retrieved from the internal radiology information system (RIS), and CXR image data was accessed via the research PACS at the \emph{Technical University of Munich (TUM) University Hospital/Klinikum rechts der Isar} (TUM UH); all records underwent de-identification before further processing.
This retrospective study was approved by the Institutional Ethics Committee of the TUM UH (approval no. 2024-93-S-CB). Owing to the retrospective design and the use of fully de-identified data, the requirement for written informed consent was waived. All procedures were performed in accordance with the Declaration of Helsinki and relevant institutional guidelines.
For dataset construction, CXR report data from $371,146$ examinations was imported from the RIS, the research PACS was then queried for corresponding DICOM files, and metadata filtering (retaining frontal AP/PA and lateral views in erect, semi-erect, and supine positions) and image quality screening were applied (further details in Supplementary Section~\ref{subsec: appendix_internal_dataset_details}). Overall, $323,562$ CXR images from $203,602$ examinations of $114,573$ patients ($55 \%$ male; mean age: $63 \pm 18$ years) acquired at TUM UH between January 2005 and March 2024 were included. Dataset splits for vision-language model training and evaluation were defined at the patient level to prevent near-duplicate leakage across splits.

\textbf{External validation dataset.} For external evaluation, we used CheXlocalize \cite{saporta2022benchmarking}, a localization-focused derivative of the CheXpert dataset, a large chest radiograph dataset from Stanford Health Care with $224{,}316$ radiographs from $65{,}240$ patients and report-derived labels for 14 observations \cite{irvin2019chexpert}.

CheXlocalize augments CheXpert validation/test data with expert localization annotations for 10 classes (Atelectasis, Cardiomegaly, Consolidation, Edema, Enlarged Cardiomediastinum, Lung Lesion, Lung Opacity, Pleural Effusion, Pneumothorax, Support Devices) \cite{saporta2022benchmarking}. The derivative dataset contains $234$ validation radiographs (from $200$ patients) and $668$ test radiographs (from $500$ patients). In addition to image-level labels, it provides radiologist-generated pixel-level segmentations; for the human-benchmark subset, it also provides additional representative point annotations \cite{saporta2022benchmarking}.

\subsection{LLM-based CXR report structuring and labeling}
\label{sec:llm-structuring}

To convert German free-text CXR reports into a uniform, machine-readable format, we defined a structured labeling template covering the clinically relevant content of a CXR report given local institution practices and requirements. The template was designed and reviewed by board-certified radiologists and comprises free-text fields (examination, clinical information, report date, residual findings, original impression) together with four hierarchical groups in which each field is represented by both an extracted text span from the original report and a categorical label: (1)~comparison information, (2)~support devices (3)~thoracic organs (heart, mediastinum), and (4)~pathologies (lung, pleural, vascular, and osseous findings). Categorical labels are binary (present/absent) for comparison and support device fields and four-class (0--3, encoding no information available, normal or absent, ambiguous or borderline, or clearly pathological) for thoracic organ and pathology fields. A complete listing of all fields and label definitions is provided in Supplementary Table~\ref{tab: appendix_field_label_definitions}.

Based on previous performance comparisons regarding information extraction from German radiology reports \cite{woznicki2024structuring,le2024performance,passweg2025data}, structuring was performed by a locally hosted endpoint of the GPT-OSS-120B large language model \cite{openai2025gptoss120bgptoss20bmodel} deployed within the hospital IT infrastructure. The model is prompted in German with task instructions that were iteratively refined on a dedicated development set ($N=100$, excluded from the potential reader study samples) in collaboration with an experienced radiologist (the full prompt is available as a translated version in the associated code repository). Outputs are constrained via guided JSON generation against a strict JSON Schema and additionally validated against a Pydantic model, with failures triggering retries at elevated reasoning effort. Implementation and inference details are given in Supplementary Section~\ref{app:llm-pipeline-details}.

To assess output quality beyond mere structural validity, we conducted a reader study with three radiology residents (varying clinical experience; supervision of two board-certified
radiologists KKB and LCA) on 200 reports drawn from the processed report pool. Sampling was predominantly random and supplemented with targeted draws covering rare field entries and low-prevalence label classes; the full sampling protocol and resulting class distribution are described in Supplementary Section~\ref{app:reader-study-sampling}, with per-field text-presence and label-class counts summarized in Supplementary Table~\ref{tab:field-screening-summary}. Each reader independently reviewed the original full-text report alongside the LLM-generated proposal for each sample individually and corrected any text spans or labels that did not faithfully reflect the report content. The reader-corrected outputs served as field-level reference: categorical fields were evaluated (pairwise reader-LLM) using macro-averaged F1 and Cohen's $\kappa$~\cite{cohen1960coefficient}, while extracted text spans were evaluated using macro BERTScore-F1~\cite{zhang2020bertscore} with contextual embeddings from the German medical-domain model medBERT.de~\cite{bressem2024medbertde}. Inter-rater agreement was additionally quantified using Krippendorff's $\alpha$~\cite{krippendorff2018content} and Gwet's AC1~\cite{gwet2008computing} for labels and pairwise (reader-LLM) BERTScore-F1 for text spans.

\subsection{RadPRISM model}

\subsubsection{Data preprocessing}
To improve rare-concept information coverage within the structured report dataset without severely distorting the empirical reporting distribution, we applied rule-based synthetic imputation at the concept level prior to alignment training. For support-device concepts, a synthetic absence description was imputed when the LLM-assigned label already indicated device absence, but the corresponding text field was empty, reflecting the common reporting practice of describing support devices only when present or recently removed. For selected pathology concepts, a synthetic absence description was imputed only for empty concept fields when the LLM-assigned label was non-pathological (or mildly/uncertain) and the surrounding pathology section showed high local coverage (more than a predefined proportion of the other pathology fields contained descriptions - 65\% in our implementation; applied rules and thresholds were defined by a board-certified radiologist based clinical reporting experience), consistent with the interpretation that the radiologist had considered the concept but omitted explicit mention because it was absent. Imputed sentences were drawn from a diverse candidate pool (Supplementary Section~\ref{subsubsec: synthetic text imputation pool}) to avoid excessive concentration in the text embedding space; only a random subset of eligible cases was filled, and the per-concept number of imputations was capped at the previously observed real-data coverage to prevent synthetic entries from dominating the dataset over spans extracted from actual CXR reports.

For the concept-wise supervised linear-head finetuning and classification evaluation, binary labels were generated by mapping support-device annotations directly to present/absent and by collapsing thoracic-organ and pathology annotations into negative (LLM-assigned label 1, normal) versus positive (LLM-assigned labels 2 and 3, ambiguous/borderline or clearly pathological), while leaving unknown states (LLM-assigned 0) unlabeled; for pathology concepts with unknown labels, negative labels were additionally imputed under the same surrounding-coverage rationale as described above for the text imputations. Concept-specific text coverage and label distributions before and after preprocessing are reported in Supplementary Tables~\ref{tab:text_label_split_overview_before_imputation} and~\ref{tab:text_label_split_overview_after_65pct_imputation}.

\subsubsection{Architecture components and concept-stratified alignment}

In the RadPRISM vision path, we employed the DINOv2-based \cite{oquab2023dinov2}, CXR-pretrained RAD-DINO-MAIRA-2 \cite{perezgarcia2024raddino,bannur2024maira2} vision transformer (ViT) image encoder, which remained frozen throughout training. For downstream processing of the ViT output image patch embeddings, we added a trainable linear projection layer that maps them to a lower-dimensional space (dimension $d_p$) for the following concept-attention layer. Following the approach by Gu et al. \cite{gu2025radalign}, this cross-attention block then introduces $K$ learnable schema-based concept tokens that aim to extract corresponding specific visual features $\mathbf{v}^{(k)}$ from the projected ViT output tokens. 

In the text path, we used Qwen3-Embedding-4B \cite{qwen3embedding} as the text embedding model for our German report scenario, owing to its strong multilingual capabilities. The text embedding model remains frozen throughout training as well. For downstream processing of the text embeddings, we also added a trainable linear projection layer in this path, which maps them to the same dimension as the concept-specific extracted visual feature vectors (dimension $d_p$). 

To then align the concept-specific extracted visual feature vectors $\mathbf{v}^{(k)}$ to the corresponding text embeddings $\mathbf{u}^{(k)}$, we use a loss function consisting of the $K$ concept-specific contrastive loss terms:

\begin{align}
\mathcal{L}
&= \frac{1}{K} \sum_{k=1}^{K} \mathcal{L}^{(k)} =\frac{1}{K} \sum_{k=1}^{K} \frac{1}{2} \left( \mathcal{L}^{(k)}_{\text{i}\rightarrow\text{t}} + \mathcal{L}^{(k)}_{\text{t}\rightarrow\text{i}} \right)
\end{align}

The concept-wise components $\mathcal{L}^{(k)}$ include symmetric image-to-text $\mathcal{L}^{(k)}_{\text{i}\rightarrow\text{t}}$ and text-to-image $\mathcal{L}^{(k)}_{\text{t}\rightarrow\text{i}}$ contrastive loss terms with concept-individual learnable temperatures $\tau_k$.
Moreover, we employed weighted sampling during batching to ensure a minimum batch coverage for concepts with particularly underrepresented text descriptions in the structured report dataset. For the individual concept-wise loss terms, only batch samples with valid text embeddings (non-empty concept descriptions) contribute to the contrastive loss; the empty text instances were masked out. 

\subsubsection{Implementation details}
\label{subsec: Implementation details}

In the used setup configuration, $K=19$ report-derived concepts were modeled, spanning support devices, thoracic organs, and pulmonary/pleural/vascular/osseous findings. The backbone embeddings were projected into a $d_p=540$-dimensional modality-shared concept space and queried by 19 learnable concept tokens through a single 12-head cross-attention block with dropout of 0.1. Precomputed 768-dimensional concept-specific text embeddings (Qwen3 embedding dimension was set to 768 for direct comparability to medBERT.de embedding ablation, see ablation section) were projected into the same space and aligned to the corresponding visual concept embeddings. Per-concept alignment temperatures were learnable, initialized at 0.07, and constrained to the range $0.02-0.12$ for stability.

For the ViT CXR image input, we used a resolution of 518 x 518 px with aspect-ratio-preserving resizing and intensity normalization as in the RAD-DINO-MAIRA-2 preprocessing. During training, we employed random affine image transformations and random image corner cut-outs (see Supplementary Section \ref{subsubsec: appendix_img_augmentations} for details) as augmentations. The image corner-cutout augmentation was used to mitigate shortcut learning by leveraging patient positioning markers/annotations typically present at the image corners (like described in DeGrave et al. \cite{degrave2021ai}). 

The full dataset was split on the patient level with a train proportion of 80 \% and validation and test of 10 \% each (for details on the split instance/patient statistics, refer to Supplementary Table \ref{tab:dataset_split_overview}. For the concept-aware weighted sampling strategy, samples containing text for concepts with less than 10 \% training-set coverage were oversampled in proportion to inverse concept coverage, with a target coverage of 10 \%, using weights derived from a precomputed concept-coverage table. Training ran for a maximum of 15 epochs (early stopping for validation loss minimum) with AdamW (learning rate $5 \cdot 10^{-5}$, weight decay $10^{-4}$), bfloat16 mixed precision, linear warm-up for the first 1,000 steps, cosine learning-rate decay thereafter, and gradient clipping at 1.0. The batch size was 256, and the final pretraining checkpoint was selected based on the lowest validation split alignment loss.
To assess setup performance variability arising from stochastic optimization, we repeated each training configuration with three different random seeds, thereby varying model initialization and stochastic training operations, and used the resulting performance spread as an estimate of run-to-run variability.

\subsection{Evaluation}
\label{sec:evaluation}

We evaluated the quality of RadPRISM concept-specific vision-language alignment via four complementary analyses: concept-wise classification (zero-shot and linear probing), visual grounding on CheXlocalize, comparison with three baseline setups, and a radiologist reader study.

\textbf{Image classification.} Zero-shot concept classification was performed using a German prompt database with three positive and three negative formulations per concept (full prompt set in Supplementary Table~\ref{tab: appendix_zero_shot_prompts}). For each concept, all positive and all negative prompts were embedded with the model's text encoder, projected into the shared concept space, normalized, and mean-pooled to obtain one positive and one negative concept prototype. The probability of concept presence was derived from a two-way positive-versus-negative similarity comparison of the concept-specific visual embedding against both prototypes, yielding one zero-shot score per concept and image without task-specific training.

Linear probing classification was performed by fine-tuning concept-wise linear heads on top of the alignment-pretrained model, with the image encoder and all other alignment parameters frozen, supervised by the LLM-derived binary labels and corresponding validity masks (label '0' = no information). Optimization used a masked binary cross-entropy loss with tempered, concept-balanced positive weights to address class imbalance; the full loss formulation, training schedule, and hyperparameters are given in Supplementary Section~\ref{app:classification-training}. Performance on the internal test split was assessed concept-wise on samples with valid labels using AUROC and AUPRC.
Uncertainty was quantified using a non-parametric percentile bootstrap. For the evaluated dataset, cases were resampled with replacement $1{,}000$ times. For each bootstrap replicate, per-concept AUROC and AUPRC were recalculated and averaged across evaluable concepts to obtain macro AUROC and macro AUPRC. The 95\% confidence intervals were defined by the 2.5th and 97.5th percentiles of the bootstrap distributions.
External generalization was evaluated on CheXlocalize~\cite{saporta2022benchmarking} using the subset of CheXpert classes mappable to our structured-report concepts (mapping in Supplementary Section~\ref{subsec: appendix_chexpert_label_class_concept_mappings}, Supplementary Table~\ref{tab:mapping_classification}); for each non-one-to-one mapped class, the maximum score across the corresponding source concepts was used.

\textbf{Visual grounding.} Visual grounding was evaluated quantitatively on CheXlocalize~\cite{saporta2022benchmarking} using concept-specific attention maps extracted from the trained concept-attention block. Only strict one-to-one concept-to-class mappings (Supplementary Section~\ref{subsec: appendix_chexpert_label_class_concept_mappings}, Supplementary Table~\ref{tab:mapping_grounding}) and cases with valid labels and ground-truth segmentations were included. The primary, threshold-independent metric was the pointing-game hit rate, where a case is counted as a hit if the pixel of maximal attention lies within the ground-truth mask. As an auxiliary overlap-based analysis, Dice and IoU were computed using concept-specific attention thresholds optimized on the CheXlocalize validation split and transferred to the test split. Image preprocessing, threshold-sweep specifics, and uncertainty estimation via $1{,}000$-sample case-level nonparametric bootstrap are described in Supplementary Section~\ref{app:grounding-details}.

\textbf{Baseline and ablation setups.} To contextualize our results, we compared against three baseline setups and one text-encoder ablation of our framework. (i) A \emph{report-level alignment} baseline that retains the architectural components of our setup (RAD-DINO-MAIRA-2 image encoder; Qwen3-Embedding-4B text embedding) but embeds the full report directly and uses a single query token in the concept-attention block, yielding a global visual feature applied to all concepts at zero-shot inference. (ii) A \emph{linear-probe} baseline that fits concept-wise linear classification heads directly on the frozen RAD-DINO-MAIRA-2 CLS token, without any vision-language pretraining, using the same LLM-derived binary labels. (iii) \emph{CARZero}~\cite{lai2024carzero}, also a cross-attention alignment-based architecture, identified by Luo et al.~\cite{luo2025xbench} as a vision-language model with leading performance for chest-radiograph classification and grounding on CheXlocalize and similar datasets. We used Luo et al.'s work-associated code implementation\footnote{\url{https://github.com/Roypic/Benchmarkingattention}} for CARZero evaluations and, for fair grounding comparison, additionally retrained RadPRISM at CARZero's native input resolution of $224\times 224$. (iv) A \emph{text-encoder ablation} of our framework that replaces the Qwen3-Embedding-4B text encoder with the German radiology-specific medBERT.de model~\cite{bressem2024medbertde}, while keeping all other components and training settings identical, isolating the contribution of the multilingual text embedding compared to a medical-language-specific model; the corresponding comparison is reported in Supplementary Section~\ref{app:qwen-medbert}. Further baseline and ablation implementation details are provided in Supplementary Section~\ref{app:baselines-details}.

\textbf{Reader study.} To complement the LLM-label-based internal evaluation and the limited mappable subset on CheXlocalize, we conducted a radiologist reader study on 100 frontal-view cases sampled from each of the internal and external test splits, balanced via concept-wise quota sampling (full sampling protocol in Supplementary Section~\ref{app:reader-study-vlm-sampling}). For each case and target concept, readers were presented with the CXR image alongside the model's attention map, binary label prediction, and the best text match(es) from a concept-stratified retrieval database, and provided structured ratings for each output. Retrieval correctness at rank 3 was defined per case as at least one of the top-3 retrieved spans being rated correct, macro-averaged across concepts.
In addition, we used the same LLM as for the structuring/labeling workflow to assess consistency between the classification prediction and the top-retrieved text (consistent, uncertain, or inconsistent = e.g., the assigned label indicates pathology present, but the retrieved text does not), and readers also provided feedback on whether this assessment was correct. A quantitative relationship analysis between consistency status and correctness rates was conducted using pairwise z-tests with Holm correction.\cite{holm1979simple}
Due to the complexity of the reader study task, each case was assessed by only one reader (with different total numbers of samples assessed per reader, depending on availability) to keep per-reader evaluation time acceptable, given the institution's annotation resources.
The full rating scheme, the derivation of per-concept operating thresholds for label binarization, and the construction of the concept-stratified retrieval database are described in Supplementary Section~\ref{app:reader-study-vlm-details}.

\section*{Author contributions}
FD, LCA, and KKB conceived and designed the study. FD designed the setups, executed the experiments, analyzed and interpreted the results, and drafted the manuscript. LS, LeSch, MK, MF, JK, and ES participated in the reader studies and contributed to data annotation. FP, JM, JL\"u, ZBC, and AN provided iterative feedback on study design, execution, and interpretation, and revised the manuscript critically for important intellectual content. FP contributed to data preprocessing. CIB, MRM, DR, and SZ contributed to the study design, supervised the study, and revised the manuscript. LCA and KKB supervised the study and revised the manuscript critically for important intellectual content. All authors reviewed and approved the final version of the manuscript and accept responsibility for the decision to submit for publication.

\section*{Competing interests}
All authors declare no competing interests.

\section*{Data availability}
The CheXlocalize dataset is publicly available at \url{https://stanford.redivis.com/datasets/efx9-5nspnbb4b} upon application for access. The in-house TUM dataset cannot be shared publicly owing to patient privacy regulations and institutional data governance requirements. Anonymized aggregate results supporting the findings of this study are available from the corresponding author on reasonable request from the date of publication. Requests will be reviewed by the study team, and a signed data access agreement will be required before release.

\section*{Code availability}
The custom code for report stratification, RadPRISM training, and evaluation is available at \url{https://github.com/FabianD191/rad_prism.git}. 

\section*{Funding}
This project was funded by the Wilhelm Sander Foundation, Munich, Germany: Grant 2025.013.1 (L.C.A, K.K.B).

\bibliographystyle{naturemag}
\bibliography{references}

@misc{openai2025gptoss120bgptoss20bmodel,
      title={gpt-oss-120b \& gpt-oss-20b Model Card}, 
      author={OpenAI},
      year={2025},
      eprint={2508.10925},
      archivePrefix={arXiv},
      primaryClass={cs.CL},
      url={https://arxiv.org/abs/2508.10925}, 
}

@inproceedings{irvin2019chexpert,
      title={CheXpert: A Large Chest Radiograph Dataset with Uncertainty Labels and Expert Comparison},
      author={Irvin, Jeremy and Rajpurkar, Pranav and Ko, Michael and Yu, Yifan and Ciurea-Ilcus, Silviana and Chute, Chris and Marklund, Henrik and Haghgoo, Behzad and Ball, Robyn and Shpanskaya, Katie and Seekins, Jayne and Mong, David A. and Halabi, Safwan and Sandberg, Jesse K. and Jones, Ricky and Larson, David B. and Langlotz, Curtis P. and Patel, Bhavik N. and Lungren, Matthew P. and Ng, Andrew Y.},
      booktitle={Proceedings of the AAAI Conference on Artificial Intelligence},
      volume={33},
      pages={590--597},
      year={2019},
      doi={10.1609/aaai.v33i01.3301590},
      url={https://ojs.aaai.org/index.php/AAAI/article/view/3834}
}

@article{saporta2022benchmarking,
      title={Benchmarking saliency methods for chest X-ray interpretation},
      author={Saporta, Adriel and Gui, Xiaotao and Agrawal, Ashwin and Pareek, Anuj and others},
      journal={Nature Machine Intelligence},
      volume={4},
      number={10},
      pages={867--878},
      year={2022},
      month={oct},
      doi={10.1038/s42256-022-00536-x},
      url={https://www.nature.com/articles/s42256-022-00536-x}
}

@inproceedings{zhang2020bertscore,
      title={BERTScore: Evaluating Text Generation with BERT},
      author={Zhang, Tianyi and Kishore, Varsha and Wu, Felix and Weinberger, Kilian Q. and Artzi, Yoav},
      booktitle={International Conference on Learning Representations},
      year={2020},
      url={https://openreview.net/forum?id=SkeHuCVFDr}
}

@article{bressem2024medbertde,
      title={medBERT.de: A German Language Model for the Medical Domain},
      author={Bressem, Keno K. and Hamm, Benjamin and Sch{"a}fer, Reinhold and Morat, Julian and Schmitt, Hannah S. and Adams, Lisa-Charlotte and Grosser, Lukas and Vahldiek, Jan and Niehues, Jan and Schlemmer, Heinz-Peter and Maier-Hein, Klaus and Kather, Jakob N. and Rueckert, Daniel and Yang, Bjoern},
      journal={Artificial Intelligence in Medicine},
      volume={157},
      pages={102928},
      year={2024},
      month={nov},
      doi={10.1016/j.artmed.2024.102928},
      url={https://www.sciencedirect.com/science/article/pii/S0933365724002130}
}

@article{cohen1960coefficient,
      title={A Coefficient of Agreement for Nominal Scales},
      author={Cohen, Jacob},
      journal={Educational and Psychological Measurement},
      volume={20},
      number={1},
      pages={37--46},
      year={1960},
      doi={10.1177/001316446002000104},
      url={https://doi.org/10.1177/001316446002000104}
}

@book{krippendorff2018content,
      title={Content Analysis: An Introduction to Its Methodology},
      author={Krippendorff, Klaus},
      edition={4},
      year={2018},
      publisher={SAGE Publications}
}

@article{gwet2008computing,
      title={Computing Inter-Rater Reliability and Its Variance in the Presence of High Agreement},
      author={Gwet, Kilem L.},
      journal={British Journal of Mathematical and Statistical Psychology},
      volume={61},
      number={1},
      pages={29--48},
      year={2008},
      doi={10.1348/000711006X126600},
      url={https://doi.org/10.1348/000711006X126600}
}

@misc{perezgarcia2024raddino,
      title={{RAD-DINO}: Exploring Scalable Medical Image Encoders Beyond Text Supervision},
      author={Fernando Pérez-García and Harshita Sharma and Sam Bond-Taylor and Kenza Bouzid and Valentina Salvatelli and Maximilian Ilse and Shruthi Bannur and Daniel C. Castro and Anton Schwaighofer and Matthew P. Lungren and Maria Wetscherek and Noel Codella and Stephanie L. Hyland and Javier Alvarez-Valle and Ozan Oktay},
      year={2024},
      eprint={2401.10815},
      archivePrefix={arXiv},
      primaryClass={cs.CV}
}

@inproceedings{gu2025radalign,
  title={Radalign: Advancing radiology report generation with vision-language concept alignment},
  author={Gu, Difei and Gao, Yunhe and Zhou, Yang and Zhou, Mu and Metaxas, Dimitris},
  booktitle={International Conference on Medical Image Computing and Computer-Assisted Intervention},
  pages={484--494},
  year={2025},
  organization={Springer}
}

@article{qwen3embedding,
  title={Qwen3 Embedding: Advancing Text Embedding and Reranking Through Foundation Models},
  author={Zhang, Yanzhao and Li, Mingxin and Long, Dingkun and Zhang, Xin and Lin, Huan and Yang, Baosong and Xie, Pengjun and Yang, An and Liu, Dayiheng and Lin, Junyang and Huang, Fei and Zhou, Jingren},
  journal={arXiv preprint arXiv:2506.05176},
  year={2025}
}

@article{degrave2021ai,
  title={AI for radiographic COVID-19 detection selects shortcuts over signal},
  author={DeGrave, Alex J and Janizek, Joseph D and Lee, Su-In},
  journal={Nature Machine Intelligence},
  volume={3},
  number={7},
  pages={610--619},
  year={2021},
  publisher={Nature Publishing Group UK London}
}

@article{luo2025xbench,
  title={XBench: A Comprehensive Benchmark for Visual-Language Explanations in Chest Radiography},
  author={Luo, Haozhe and Shu, Shelley Zixin and Zhou, Ziyu and Otalora, Sebastian and Reyes, Mauricio},
  journal={arXiv preprint arXiv:2510.19599},
  year={2025}
}

@inproceedings{lai2024carzero,
  title={Carzero: Cross-attention alignment for radiology zero-shot classification},
  author={Lai, Haoran and Yao, Qingsong and Jiang, Zihang and Wang, Rongsheng and He, Zhiyang and Tao, Xiaodong and Zhou, S Kevin},
  booktitle={Proceedings of the IEEE/CVF Conference on Computer Vision and Pattern Recognition},
  pages={11137--11146},
  year={2024}
}

@article{passweg2025data,
  title={Data Extraction from Oncology Imaging Reports by Large Language Models: A Comparative Accuracy Study},
  author={Passweg, Lea P and Schwenke, Johannes M and Schoenenberger, Christof M and Locher, Flavio and Picker, Julia and Dieterle, Manuel and Thiele, Benjamin and Hasler, Dimitri and Danelli, Alessia and Schmitt, Andreas M and others},
  journal={medRxiv},
  pages={2025--12},
  year={2025},
  publisher={Cold Spring Harbor Laboratory Press}
}

@article{le2024performance,
  title={Performance of an open-source large language model in extracting information from free-text radiology reports},
  author={Le Guellec, Bastien and Lef{\`e}vre, Alexandre and Geay, Charlotte and Shorten, Lucas and Bruge, Cyril and Hacein-Bey, Lotfi and Amouyel, Philippe and Pruvo, Jean-Pierre and Kuchcinski, Gregory and Hamroun, Aghiles},
  journal={Radiology: Artificial Intelligence},
  volume={6},
  number={4},
  pages={e230364},
  year={2024},
  publisher={Radiological Society of North America}
}

@inproceedings{zhang2022contrastive,
  title={Contrastive learning of medical visual representations from paired images and text},
  author={Zhang, Yuhao and Jiang, Hang and Miura, Yasuhide and Manning, Christopher D and Langlotz, Curtis P},
  booktitle={Machine learning for healthcare conference},
  pages={2--25},
  year={2022},
  organization={PMLR}
}

@article{park2026radzero,
  title={RadZero: Similarity-Based Cross-Attention for Explainable Vision-Language Alignment in Chest X-ray with Zero-Shot Multi-Task Capability},
  author={Park, Jonggwon and Yoon, Byungmu and Kim, Soobum and Choi, Kyoyun},
  journal={Advances in Neural Information Processing Systems},
  volume={38},
  pages={56008--56034},
  year={2026}
}

@inproceedings{gu2026anatomy,
  title={Anatomy-VLM: A fine-grained vision-language model for medical interpretation},
  author={Gu, Difei and Gao, Yunhe and Zhou, Mu and Metaxas, Dimitris},
  booktitle={Proceedings of the IEEE/CVF Winter Conference on Applications of Computer Vision},
  pages={2838--2847},
  year={2026}
}

@article{bannur2024maira2,
  title={Maira-2: Grounded radiology report generation},
  author={Bannur, Shruthi and Bouzid, Kenza and Castro, Daniel C and Schwaighofer, Anton and Thieme, Anja and Bond-Taylor, Sam and Ilse, Maximilian and P{\'e}rez-Garc{\'\i}a, Fernando and Salvatelli, Valentina and Sharma, Harshita and others},
  journal={arXiv preprint arXiv:2406.04449},
  year={2024}
}

@article{oquab2023dinov2,
  title={Dinov2: Learning robust visual features without supervision},
  author={Oquab, Maxime and Darcet, Timoth{\'e}e and Moutakanni, Th{\'e}o and Vo, Huy and Szafraniec, Marc and Khalidov, Vasil and Fernandez, Pierre and Haziza, Daniel and Massa, Francisco and El-Nouby, Alaaeldin and others},
  journal={arXiv preprint arXiv:2304.07193},
  year={2023}
}

@article{tiu2022chexzero,
  title={Expert-level detection of pathologies from unannotated chest X-ray images via self-supervised learning},
  author={Tiu, Ekin and Talius, Ellie and Patel, Pujan and Langlotz, Curtis P. and Ng, Andrew Y. and Rajpurkar, Pranav},
  journal={Nature Biomedical Engineering},
  volume={6},
  number={12},
  pages={1399--1406},
  year={2022},
  doi={10.1038/s41551-022-00936-9}
}

@article{hamamci2024ctrate,
  title={Generalist foundation models from a multimodal dataset for 3D computed tomography},
  author={Hamamci, Ibrahim Ethem and Er, Sezgin and Wang, Chenyu and Almas, Furkan and Simsek, Ayse Gulnihan and Esirgun, Sevval Nil and Dogan, Irem and Durugol, Omer Faruk and Hou, Benjamin and Shit, Suprosanna and others},
  journal={Nature Biomedical Engineering},
  pages={1--19},
  year={2026},
  publisher={Nature Publishing Group UK London}
}

@inproceedings{huang2021gloria,
  title={{GLoRIA}: A Multimodal Global-Local Representation Learning Framework for Label-efficient Medical Image Recognition},
  author={Huang, Shih-Cheng and Shen, Liyue and Lungren, Matthew P. and Yeung, Serena},
  booktitle={Proceedings of the IEEE/CVF International Conference on Computer Vision (ICCV)},
  pages={3942--3951},
  year={2021}
}

@inproceedings{boecking2022biovil,
  title={Making the most of text semantics to improve biomedical vision--language processing},
  author={Boecking, Benedikt and Usuyama, Naoto and Bannur, Shruthi and Castro, Daniel C. and Schwaighofer, Anton and Hyland, Stephanie and Wetscherek, Maria and Naumann, Tristan and Nori, Aditya and Alvarez-Valle, Javier and Poon, Hoifung and Oktay, Ozan},
  booktitle={European Conference on Computer Vision (ECCV)},
  pages={1--21},
  year={2022}
}

@inproceedings{wu2023medklip,
  title={{MedKLIP}: Medical Knowledge Enhanced Language-Image Pre-Training for {X-ray} Diagnosis},
  author={Wu, Chaoyi and Zhang, Xiaoman and Zhang, Ya and Wang, Yanfeng and Xie, Weidi},
  booktitle={Proceedings of the IEEE/CVF International Conference on Computer Vision (ICCV)},
  year={2023}
}

@article{zhang2023kad,
  title={Knowledge-enhanced visual--language pre-training on chest radiology images},
  author={Zhang, Xiaoman and Wu, Chaoyi and Zhang, Ya and Xie, Weidi and Wang, Yanfeng},
  journal={Nature Communications},
  volume={14},
  pages={4542},
  year={2023},
  doi={10.1038/s41467-023-40260-7}
}

@inproceedings{cheng2026rethinking,
  title={Rethinking radiology report generation: From narrative flow to topic-guided findings},
  author={Cheng, Sheng and Subramanian, Devika},
  booktitle={The Fourteenth International Conference on Learning Representations},
  year={2026}
}

@article{krippendorff2011computing,
  title={Computing Krippendorff's alpha-reliability},
  author={Krippendorff, Klaus},
  journal={Departmental Papers (ASC), University of Pennsylvania},
  year={2011}
}

@article{lin2024ctglip,
  title={Ct-glip: 3d grounded language-image pretraining with ct scans and radiology reports for full-body scenarios},
  author={Lin, Jingyang and Xia, Yingda and Zhang, Jianpeng and Yan, Ke and Cao, Kai and Lu, Le and Luo, Jiebo and Zhang, Ling},
  journal={arXiv preprint arXiv:2404.15272},
  year={2024}
}

@article{woznicki2024structuring,
  title={Automatic structuring of radiology reports with on-premise open-source large language models},
  author={Wo{\'z}nicki, Piotr and Laqua, Caroline and Fiku, Ina and Hekalo, Amar and Truhn, Daniel and Engelhardt, Sandy and Kather, Jakob N. and Foersch, Sebastian and Akinci D'Antonoli, Tugba and Pinto Dos Santos, Daniel and Bae{\ss}ler, Bettina and Laqua, Fabian Christopher},
  journal={European Radiology},
  volume={35},
  number={4},
  pages={2018--2029},
  year={2025},
  doi={10.1007/s00330-024-11074-y}
}

@article{imran2026multimodal,
  title={Multimodal Vision--Language Models in Medical Imaging: A Survey of Retrieval, Interpretability, and Trust},
  author={Imran, Muhammad and Lee, Yugyung},
  journal={IEEE Access},
  year={2026},
  publisher={IEEE}
}

@article{holm1979simple,
  title={A simple sequentially rejective multiple test procedure},
  author={Holm, Sture},
  journal={Scandinavian journal of statistics},
  pages={65--70},
  year={1979},
  publisher={JSTOR}
}

@inproceedings{yuksekgonul2023when,
  title     = {When and Why Vision-Language Models Behave like Bags-Of-Words, and What to Do About It?},
  author    = {Yuksekgonul, Mert and Bianchi, Federico and Kalluri, Pratyusha and Jurafsky, Dan and Zou, James},
  booktitle = {International Conference on Learning Representations (ICLR)},
  year      = {2023},
  url       = {https://openreview.net/forum?id=KRLUvxh8uaX}
}

@inproceedings{parcalabescu2022valse,
  title     = {{VALSE}: A Task-Independent Benchmark for Vision and Language Models Centered on Linguistic Phenomena},
  author    = {Parcalabescu, Letitia and Cafagna, Michele and Muradjan, Lilitta and Frank, Anette and Calixto, Iacer and Gatt, Albert},
  booktitle = {Proceedings of the 60th Annual Meeting of the Association for Computational Linguistics (ACL)},
  pages     = {8253--8280},
  year      = {2022},
  doi       = {10.18653/v1/2022.acl-long.567}
}

@article{moor2023generalist,
  title   = {Foundation models for generalist medical artificial intelligence},
  author  = {Moor, Michael and Banerjee, Oishi and Abad, Zahra Shakeri Hossein and Krumholz, Harlan M. and Leskovec, Jure and Topol, Eric J. and Rajpurkar, Pranav},
  journal = {Nature},
  volume  = {616},
  number  = {7956},
  pages   = {259--265},
  year    = {2023},
  doi     = {10.1038/s41586-023-05881-4}
}

@article{christensen2024vision,
  title   = {Vision-language foundation model for echocardiogram interpretation},
  author  = {Christensen, Matthew and Vukadinovic, Milos and Yuan, Neal and Ouyang, David},
  journal = {Nature Medicine},
  volume  = {30},
  number  = {5},
  pages   = {1481--1488},
  year    = {2024},
  doi     = {10.1038/s41591-024-02959-y}
}

@inproceedings{miura2021improving,
  title     = {Improving Factual Completeness and Consistency of Image-to-Text Radiology Report Generation},
  author    = {Miura, Yasuhide and Zhang, Yuhao and Tsai, Emily Bao and Langlotz, Curtis P. and Jurafsky, Dan},
  booktitle = {Proceedings of the 2021 Conference of the North American Chapter of the Association for Computational Linguistics: Human Language Technologies (NAACL-HLT)},
  pages     = {5288--5304},
  year      = {2021},
  doi       = {10.18653/v1/2021.naacl-main.416}
}

@article{yu2023evaluating,
  title   = {Evaluating progress in automatic chest X-ray radiology report generation},
  author  = {Yu, Feiyang and Endo, Mark and Krishnan, Rayan and Pan, Ian and Tsai, Andy and Reis, Eduardo Pontes and Fonseca, Eduardo Kaiser Ururahy Nunes and Lee, Henrique Min Ho and Abad, Zahra Shakeri Hossein and Ng, Andrew Y. and Langlotz, Curtis P. and Venugopal, Vasantha Kumar and Rajpurkar, Pranav},
  journal = {Patterns},
  volume  = {4},
  number  = {9},
  pages   = {100802},
  year    = {2023},
  doi     = {10.1016/j.patter.2023.100802}
}

@article{abdar2021review,
  title   = {A review of uncertainty quantification in deep learning: Techniques, applications and challenges},
  author  = {Abdar, Moloud and Pourpanah, Farhad and Hussain, Sadiq and Rezazadegan, Dana and Liu, Li and Ghavamzadeh, Mohammad and Fieguth, Paul and Cao, Xiaochun and Khosravi, Abbas and Acharya, U. Rajendra and Makarenkov, Vladimir and Nahavandi, Saeid},
  journal = {Information Fusion},
  volume  = {76},
  pages   = {243--297},
  year    = {2021},
  doi     = {10.1016/j.inffus.2021.05.008}
}

@article{kompa2021second,
  title   = {Second opinion needed: communicating uncertainty in medical machine learning},
  author  = {Kompa, Benjamin and Snoek, Jasper and Beam, Andrew L.},
  journal = {npj Digital Medicine},
  volume  = {4},
  number  = {1},
  pages   = {4},
  year    = {2021},
  doi     = {10.1038/s41746-020-00367-3}
}

@article{ghassemi2021false,
  title   = {The false hope of current approaches to explainable artificial intelligence in health care},
  author  = {Ghassemi, Marzyeh and Oakden-Rayner, Luke and Beam, Andrew L.},
  journal = {The Lancet Digital Health},
  volume  = {3},
  number  = {11},
  pages   = {e745--e750},
  year    = {2021},
  doi     = {10.1016/S2589-7500(21)00208-9}
}

@article{reyes2020interpretability,
  title   = {On the Interpretability of Artificial Intelligence in Radiology: Challenges and Opportunities},
  author  = {Reyes, Mauricio and Meier, Raphael and Pereira, S{\'e}rgio and Silva, Carlos A. and Dahlweid, Fried-Michael and von Tengg-Kobligk, Hendrik and Summers, Ronald M. and Wiest, Roland},
  journal = {Radiology: Artificial Intelligence},
  volume  = {2},
  number  = {3},
  pages   = {e190043},
  year    = {2020},
  doi     = {10.1148/ryai.2020190043}
}

@article{willemink2020preparing,
  title   = {Preparing Medical Imaging Data for Machine Learning},
  author  = {Willemink, Martin J. and Koszek, Wojciech A. and Hardell, Cailin and Wu, Jie and Fleischmann, Dominik and Harvey, Hugh and Folio, Les R. and Summers, Ronald M. and Rubin, Daniel L. and Lungren, Matthew P.},
  journal = {Radiology},
  volume  = {295},
  number  = {1},
  pages   = {4--15},
  year    = {2020},
  doi     = {10.1148/radiol.2020192224}
}

@article{varoquaux2022machine,
  title   = {Machine learning for medical imaging: methodological failures and recommendations for the future},
  author  = {Varoquaux, Ga{\"e}l and Cheplygina, Veronika},
  journal = {npj Digital Medicine},
  volume  = {5},
  number  = {1},
  pages   = {48},
  year    = {2022},
  doi     = {10.1038/s41746-022-00592-y}
}

@article{muehlematter2021approval,
  title   = {Approval of artificial intelligence and machine learning-based medical devices in the {USA} and {Europe} (2015--20): a comparative analysis},
  author  = {Muehlematter, Urs J. and Daniore, Paola and Vokinger, Kerstin N.},
  journal = {The Lancet Digital Health},
  volume  = {3},
  number  = {3},
  pages   = {e195--e203},
  year    = {2021},
  doi     = {10.1016/S2589-7500(20)30292-2}
}

\ifappendixatend
\clearpage
\onecolumn
\setcounter{secnumdepth}{3}   % number the Supplementary sections again
\setcounter{section}{0}
\setcounter{subsection}{0}
\setcounter{subsubsection}{0}
\setcounter{table}{0}
\setcounter{figure}{0}
\renewcommand{\thefigure}{\arabic{figure}}
\renewcommand{\thetable}{\arabic{table}}
\renewcommand{\figurename}{Supplementary Fig.}
\renewcommand{\tablename}{Supplementary Table}
\captionsetup{width=0.9\textwidth}

\section{Supplementary Information}
\subsection{Supplementary methods}

\subsubsection{Structuring and labeling schema definitions}

\begin{table}[htpb]
  \centering
  \small
  \setlength{\tabcolsep}{4.5pt}
  \renewcommand{\arraystretch}{1.15}
  \caption{Label schema and field definitions used for CXR report structuring. Label value binary means possible entries are 'existent'/'not existent'.}
  \begin{tabular}{p{0.30\linewidth} l p{0.42\linewidth}}
    \toprule
    Field / subfield & Label values & Descriptor \\
    \midrule
    Examination & -- & Name of the examination. \\
    Clinical information & -- & Indication/question and clinical information. \\
    Report date & -- & Date of the current report. \\
    Comparison & binary & Comparison study referenced (and if present, its date and modality). \\
    \textbf{Support devices} & & \\
    \quad Airway & binary & Airway support devices (e.g., endotracheal tube, tracheostomy cannula). \\
    \quad Feeding tube & binary & Gastric/enteric tube (oro-/nasogastric, etc.). \\
    \quad Central venous access & binary & Central venous catheter (CVC), port, Hickman, PICC, etc. \\
    \quad Chest drain & binary & Thoracic drain/chest tube. \\
    \quad Pacemaker & binary & Pacemaker/ICD/CRT-D, etc. \\
    \quad Other & binary & Other devices/support (ECMO, Impella, IABP, clips, cerclage wires, etc.). \\
    Heart & 0/1/2/3 & Heart status. \\
    Mediastinum & 0/1/2/3 & Mediastinum/hila/aorta status. \\
    \textbf{Pathologies} & & \\
    \quad Lung: Pneumonia & 0/1/2/3 & Pneumonia/infiltrates/opacities/consolidation. \\
    \quad Lung: Atelectasis & 0/1/2/3 & Ventilatory disturbance/atelectasis. \\
    \quad Lung: Emphysema & 0/1/2/3 & Pulmonary emphysema (COPD). \\
    \quad Lung: Fibrosis & 0/1/2/3 & Fibrosis/interstitial markings increase (non-cardiogenic). \\
    \quad Lung: Mass lesion & 0/1/2/3 & Intrapulmonary mass/nodule/tumor suspicion (not effusions, atelectasis, edema, nonspecific opacities without nodule character). \\
    \quad Pleura: Pneumothorax & 0/1/2/3 & Pneumothorax status. \\
    \quad Pleura: Pleural effusion & 0/1/2/3 & Pleural effusion/blunting of the costophrenic angle. \\
    \quad Vessels: Congestion & 0/1/2/3 & Pulmonary vascular congestion/cephalisation/redistribution, etc. \\
    \quad Vessels: Pulmonary edema & 0/1/2/3 & Interstitial/alveolar oedema. \\
    \quad Bones: Fracture & 0/1/2/3 & Fracture(s). \\
    \quad Bones: Other & 0/1/2/3 & Other bony findings (e.g., osteopenia, degenerative changes, scoliosis). \\
    Other findings & -- & Residual findings without a dedicated field (e.g., soft-tissue emphysema, elevated hemidiaphragm, projection artifacts). \\
    Impression & -- & Original impression text. \\
    \bottomrule
  \end{tabular}
  \label{tab: appendix_field_label_definitions}
\end{table}

\subsubsection{LLM structuring and labeling pipeline -- implementation details}
\label{app:llm-pipeline-details}

For each report, the structuring pipeline (cf.\ the main-text Methods, \nameref{sec:llm-structuring}) constructs a two-message chat prompt for the locally hosted GPT-OSS-120B endpoint~\cite{openai2025gptoss120bgptoss20bmodel}: (i)~a system message containing the full task instructions (see associated code repository), and (ii)~a user message passing the report content as plain text, with the examination name and the full-text report body explicitly separated.

Inference is configured for quasi-deterministic output using \texttt{temperature=0} and \texttt{top\_p=1}, a maximum of $4{,}000$ output tokens, and a reasoning effort of \texttt{low} on the initial call, increased to \texttt{medium} on validation retries. Structured output is enforced via guided JSON generation against a strict JSON Schema reflecting the target template described in the main text. Each output undergoes two-stage validation: (1)~JSON parsing and (2)~schema validation against a Pydantic model configured to reject unexpected keys and enforce field-level constraints. On validation failure, a corrective user message instructing the model to return valid JSON is appended, and the request is retried with increased reasoning effort, continuing until validation succeeds or the retry limit is reached.

\begin{table}[htpb]
  \centering
  \small
  \setlength{\tabcolsep}{4.5pt}
  \renewcommand{\arraystretch}{1.15}
  \caption{Structured report field screening summary (N=200). For each field/subfield: text presence counts (present/empty). For binary labels: present/not present. For ordinal labels: counts for 0/1/2/3. Fields without labels are marked with --.}
  \label{tab:field-screening-summary}

  \begin{tabular}{p{0.33\linewidth} cc cc cccc}
    \toprule
    & \multicolumn{2}{c}{Text} & \multicolumn{2}{c}{Binary label} & \multicolumn{4}{c}{Ordinal label} \\
    \cmidrule(lr){2-3}\cmidrule(lr){4-5}\cmidrule(lr){6-9}
    Field / subfield & present & empty & present & not present & 0 & 1 & 2 & 3 \\
    \midrule
    Examination & 200 & 0 & -- & -- & -- & -- & -- & -- \\
    Clinical information & 200 & 0 & -- & -- & -- & -- & -- & -- \\
    Report date & 0 & 200 & -- & -- & -- & -- & -- & -- \\
    Comparison & 107 & 93 & 107 & 93 & -- & -- & -- & -- \\
    \textbf{Support devices} & & & & & & & & \\
    \quad Airway & 16 & 184 & 8 & 192 & -- & -- & -- & -- \\
    \quad Feeding tube & 14 & 186 & 12 & 188 & -- & -- & -- & -- \\
    \quad Central venous access & 39 & 161 & 36 & 164 & -- & -- & -- & -- \\
    \quad Chest drain & 17 & 183 & 12 & 188 & -- & -- & -- & -- \\
    \quad Pacemaker & 17 & 183 & 16 & 184 & -- & -- & -- & -- \\
    \quad Other & 15 & 185 & 14 & 186 & -- & -- & -- & -- \\
    Heart & 151 & 49 & -- & -- & 48 & 90 & 18 & 44 \\
    Mediastinum & 134 & 66 & -- & -- & 66 & 81 & 9 & 44 \\
    \textbf{Pathologies} & & & & & & & & \\
    \quad Lung: Pneumonia & 180 & 20 & -- & -- & 20 & 143 & 27 & 10 \\
    \quad Lung: Atelectasis & 42 & 158 & -- & -- & 158 & 12 & 11 & 19 \\
    \quad Lung: Emphysema & 32 & 168 & -- & -- & 168 & 5 & 8 & 19 \\
    \quad Lung: Fibrosis & 15 & 185 & -- & -- & 185 & 5 & 4 & 6 \\
    \quad Lung: Mass lesion & 28 & 172 & -- & -- & 172 & 18 & 3 & 7 \\
    \quad Pleura: Pneumothorax & 123 & 77 & -- & -- & 77 & 111 & 4 & 8 \\
    \quad Pleura: Pleural effusion & 178 & 22 & -- & -- & 22 & 127 & 12 & 39 \\
    \quad Vessels: Congestion & 156 & 44 & -- & -- & 44 & 122 & 13 & 21 \\
    \quad Vessels: Pulmonary edema & 13 & 187 & -- & -- & 187 & 5 & 4 & 4 \\
    \quad Bones: Fracture & 45 & 155 & -- & -- & 155 & 33 & -- & 12 \\
    \quad Bones: Other & 76 & 124 & -- & -- & 125 & 4 & 53 & 18 \\
    Other findings & 149 & 51 & -- & -- & -- & -- & -- & -- \\
    Impression & 200 & 0 & -- & -- & -- & -- & -- & -- \\
    \bottomrule
  \end{tabular}

  \label{tab: appendix_structuring_reader_study_samples_overview}
\end{table}

\subsubsection{Reader study sampling protocol (Structuring/Labeling evaluation)}
\label{app:reader-study-sampling}

To balance comprehensive evaluation with a manageable reader workload, 200 samples were selected from the processed report pool, excluding the development set. Sampling was predominantly random but included a targeted minority of low-prevalence cases to ensure coverage of rare field entries. Specifically, 10 reports were sampled for each of three support device subfields with low text-entry prevalence (airway devices, feeding tube, pacemaker), and 4 reports per non-zero label class (labels 1, 2, and 3) were sampled for three low-prevalence pathology fields (lung fibrosis, lung emphysema, pulmonary edema), yielding 66 targeted samples in total. The remaining 134 samples were drawn at random. Label and text statistics for the full sample pool are reported in Supplementary Table~\ref{tab: appendix_structuring_reader_study_samples_overview}.

\subsubsection{Synthetic text imputation pool}
\label{subsubsec: synthetic text imputation pool}

In the following, the used synthetic text sampling pools for the selected concepts is defined (English translation from the German sentences):

\bigskip
\noindent\textbf{Support devices: Airway}\\
No tube in place.
No endotracheal tube present.
Airways clear, no devices in place.
Free airways without tube or tracheostomy cannula.
No evidence of airway devices in place.
Trachea freely patent, no tube.
No tracheal intubation.
Neither tube nor tracheostomy cannula in place.
No airway device identifiable.
Trachea and main bronchi clear, no therapeutic devices.
Without airway device in place.
No evidence of endotracheal intubation.
Airways without material in place.
No airway therapeutic devices delineable.
No ventilation tube detectable.
No orally or nasally inserted tube visible.
No foreign material along the course of the trachea.
No tracheostomy cannula identifiable.
No evidence of a tracheostomy cannula.
No endoluminal material in the trachea.
Airway system free of therapeutic material.
No evidence of endotracheal or tracheostomic airway management.
No foreign material in the region of the central airways.
No signs of an invasive airway device.
No device identifiable in the trachea.
Airway without radiologically visible devices.
No airway management device detectable.
No tube in place is seen.
No artificial airway device in the field of view.
No endotracheal or tracheostomic intervention visible.
Central airways clear, without devices.
No cannula and no tube along the tracheal course.
No airway therapeutic device identifiable in the thorax.
Without evidence of a secured artificial airway.
No material for invasive ventilation visible.
Neither endotracheal nor tracheal device detectable.
No tracheobronchial therapeutic device in place.

\bigskip
\noindent\textbf{Support devices: Pacemaker}\\
No pacemaker in place.
No cardiac pacemaker or ICD detectable.
No implantable cardiac device.
No pacemaker generator delineable.
No pacemaker lead identifiable.
No evidence of pacemaker or defibrillator.
No pacemaker.
No implanted pacemaker system.
Neither pacemaker nor ICD in place.
No generator in the region of the chest wall.
No pacemaker leads in their course.
No evidence of a cardiac pacemaker.
Thorax without implanted pacemaker system.
No cardiac generator demonstrable.
No evidence of a CRT system.
No electrodes of a cardiac stimulation system visible.
No defibrillator generator projected.
No intracardiac leads of a pacemaker.
No implanted rhythm device detectable.
No cardiac stimulation system delineable.
Neither pacemaker nor ICD or CRT present.
No pectoral cardiac generator.
No transvenous pacemaker electrodes identifiable.
No evidence of an implantable cardioverter or defibrillator.
No stimulation lead demonstrable in the right heart.
No subcutaneously projected pacemaker generator.
No cardiac device therapy radiologically identifiable.
Chest radiograph without signs of a pacemaker system.
No implanted defibrillator in the thoracic region.
No lead position consistent with a pacemaker.
No pacemaker system detectable.
No pacemaker generator demonstrable.
No biventricular resynchronisation system visible.
No evidence of cardiac electrotherapy.
No evidence of a generator with transvenous electrodes.
No leads along the course of the superior vena cava and right heart.
No cardiac device for rhythm therapy identifiable.
No ICD electrode detectable.
Neither generator nor intracardiac leads visible.
No implanted lead for cardiac stimulation.
No thoracically projected cardiac pacemaker generator.
No pacemaker or ICD system is seen.

\bigskip
\noindent\textbf{Support devices: Feeding tube}\\
No gastric tube in place.
No evidence of a nasogastric tube.
No gastrointestinal tube detectable.
No tube material in place in the oesophagus.
No enteral tube delineable.
No evidence of a gastric tube in place.
Neither gastric nor duodenal tube in place.
No nasogastric tube material identifiable.
Oesophagus clear, no tube in place.
No tube along the course of the upper GI tract.
No nasogastric or nasoduodenal tube in place.
No tubing material via the oesophagus into the stomach.
No evidence of an enteral feeding tube.
No feeding tube.
No oesophagogastric tube demonstrable.
No gastric tube position detectable.
Neither nasogastric nor nasojejunal tube visible.
No evidence of a feeding tube.
No enteral tube in the thoracic or upper abdominal region.
Without feeding tube in place.
No radiological evidence of a gastric tube.
No duodenal or jejunal feeding tube delineable.
No foreign material along the course of the oesophagus and stomach consistent with a tube.
No enteral tube in the field of view.
No evidence of a gastric drainage tube.
Neither feeding tube nor enteral drainage tube visible.
No evidence of a nasoenteral tube in place.
Oesophagus and stomach without tube in place.

\bigskip
\noindent\textbf{Support devices: Other}\\
No additional devices identifiable.
No evidence of other therapeutic devices.
No signs of further therapeutic devices.
No other therapeutic aids delineable.
No further medical devices visible.
Beyond that, no auxiliary material in place.
No additional tubes or catheters detectable.
No further foreign material in the thoracic region.
No other implanted or inserted systems identifiable.
Apart from the described findings, no further therapeutic devices.
No additional therapeutic foreign materials demonstrable.
No further device in the field of view.
No other medical-technical devices projected.
No evidence of additional thoracic devices.
Beyond that, no medical materials in place.
No other tubes, probes, or drains identifiable.
No further invasive devices delineable.
Otherwise no therapeutic devices in the thorax.
Without evidence of other thoracic devices.
No further catheter or tubing systems in the field of view.
No additional therapeutic implants are seen.
No further devices in place.
No other device projecting onto the thorax.
No additional medical systems detectable.
Otherwise no therapeutic devices radiologically visible.
No evidence of further therapeutic materials.
Thorax without other devices.
No further therapeutically relevant foreign materials.
No evidence of further medical implants or access lines.
Beyond the mentioned findings, no foreign material.
No other thoracically projected therapeutic devices.
No further medical-technical material in the field of view.
Without additional catheters, probes, or generators.
No further therapeutic foreign material is found.
No further interventional material identifiable.
Chest radiograph without further therapeutic devices.
No evidence of other devices in place.

\bigskip
\noindent\textbf{Support devices: Chest drain}\\
No chest drain in place.
No evidence of a pleural drain.
No drainage in the pleural space.
No drainage material in place.
No evidence of a chest drain in place.
No pleural drain delineable.
No thoracic drainage system identifiable.
Pleural space without drainage in place.
No evidence of a pleural drain bilaterally.
No chest drain on the left or right.
Without pleural drainage placement.
No thoracic suction drain identifiable.
Neither on the right nor on the left drainage material in the thorax.
No evidence of an intrathoracic drain.
Pleural cavities free of drainage material.
No thoracic drainage placement.
No pleural drainage catheter identifiable.
No interventional drainage material in the thorax.
No chest drain in bilateral comparison.
No evidence of a left- or right-sided pleural drain.
No drain with intrapleural position demonstrable.
No evidence of a chest drain in place.
No pleural drainage material is found.
Thorax without drainage catheter.
No evidence of a thoracic drainage tube.
No intrapleural drainage placement delineable.
No drainage material in place in the pleural space.

\bigskip
\noindent\textbf{Support devices: Central venous access}\\
No central venous catheter in place.
A central venous catheter is not in place.
No CVC detectable.
A CVC is not detectable.
No central venous catheters identifiable.
No evidence of a central venous catheter.
Neither CVC nor port catheter present.
No central venous access identifiable.
No venous access lines in the thoracic region.
No central vascular access delineable.
Neither central venous catheter nor port or PICC identifiable.
No transvenously coursing catheter placement.
No intrathoracic projection of a venous access.
No implanted venous port system.
No central venous catheter system in the field of view.
Thorax without CVC or port system.
No radiological evidence of a venous port system.
No central venous access is seen.
No central venous access detectable.
No central venous catheter placement identifiable.
No central venous access demonstrable.
No central venous access placement delineable.
No central venous access is found.
No signs of central venous access.
No central venous catheter position visible.
No central venous access placement in the field of view.
No material of a central venous access in place.
No central venous catheter material detectable.
No central venous foreign material identifiable.
No catheter placement consistent with central venous access.
No depiction of a central venous access.

\bigskip
\noindent\textbf{Pathologies -- Vessels: Pulmonary edema}\\
No pulmonary oedema.
No evidence of pulmonary oedema.
No interstitial or alveolar oedema.
Lungs bilaterally without oedema.
No radiological evidence of cardiogenic pulmonary oedema.
No signs of pulmonary venous congestion with oedema.
No interstitial congestion oedema detectable.
No alveolar opacities consistent with oedema.
No evidence of alveolar pulmonary oedema.
No perihilar oedema opacities.
Lung parenchyma without oedema-typical densities.
No picture of florid pulmonary oedema.
No diffuse alveolar reduced transparency consistent with oedema.
No evidence of cardiac congestion lung.
No confluent perihilar infiltrates consistent with oedema.
No evidence of acute pulmonary congestion.
No interstitial densities compatible with oedema.
No evidence of an alveolar-interstitial oedema pattern.
No signs of oedema in both lungs.
No evidence of congestion lung.
No symmetrical alveolar infiltrates consistent with oedema.
Lungs without signs of interstitial fluid congestion.
No evidence of oedema-related parenchymal opacification.
No florid pulmonary congestion oedema.
No diffuse interstitial markings increase consistent with oedema.
No acute cardiac pulmonary oedema demonstrable.
No evidence of alveolar fluid accumulation.
Morphologically no pulmonary oedema.
No radiological criteria for pulmonary oedema.
No interstitial-alveolar congestion pattern.
No signs of congestion with development of pulmonary oedema.
Lung without evidence of oedema-typical parenchymal densities.
No evidence of pulmonary venous oedema.
No signs of pulmonary oedema are found.
No oedema-suspicious parenchymal finding.

\bigskip
\noindent\textbf{Pathologies -- Bones: Fracture}\\
No fracture detectable.
No evidence of osseous injury.
No rib fracture identifiable.
No evidence of a fracture in the thoracic region.
Bony thorax, as far as assessable, intact.
Skeleton without signs of fracture.
Ribs bilaterally without evidence of fracture.
No osseous destructions or fractures.
Bony thoracic skeleton age-appropriately unremarkable.
No signs of fracture in the visible skeletal region.
Ribs and vertebral bodies without pathological fracture.
No evidence of acute osseous sequelae of trauma.
No fresh osseous lesion demonstrable.
No cortical disruption in the imaged thoracic skeleton.
No evidence of traumatic bone injuries.
Visible bony structures without fracture line.
No fracture lines at ribs, clavicles, or scapulae.
No evidence of osseous discontinuity.
No acute fracture in the imaged skeleton.
Osseous thorax without evidence of traumatic lesion.
No evidence of a fresh serial rib fracture.
As far as assessable, no osseous correlate of trauma.
No evidence of fracture at the depicted ribs.
Sternum, clavicles, and ribs without definite signs of fracture.
No radiological signs of osseous thoracic injury.
No evidence of a pathological or traumatic fracture.
The visible osseous structures are unremarkable with respect to fracture.
No indication of a fresh cortical disruption.
No evidence of an osseous lesion with fracture character.
No post-traumatic signs of fracture.
Thoracic skeleton without acute signs of injury.
No evidence of injury to ribs or shoulder girdle.
Visible skeleton without indication of discontinuity.
No evidence of an acute rib or clavicle fracture.
No acute fractures of the depicted bony structures.
Without radiological evidence of fracture.
No osseous trauma identifiable.
No fracture in the region of the visible thoracic skeleton.
No signs of fracture are seen.
No osseous cortical disruption detectable.

\bigskip
\noindent\textbf{Pathologies -- Lung: Emphysema}\\
No emphysema.
No evidence of pulmonary emphysema.
No emphysematous changes.
No evidence of hyperinflation of the lung parenchyma.
Lung without signs of emphysema.
No rarefaction of the pulmonary vascular markings.
No evidence of chronic hyperinflation.
Lung parenchyma without emphysematous destruction.
No bullous changes.
Lungs bilaterally without signs of emphysema.
No signs of obstructive hyperinflation.
No evidence of emphysematous areas.
Lung structure unremarkable, no emphysema.
No emphysema-typical increased transparency.
Diaphragm position normal, no evidence of hyperinflation.
No radiological indication of COPD-typical hyperinflation.
No bullous emphysema detectable.
Lung fields without emphysema-typical increased transparency.
No flattening of the diaphragm consistent with hyperinflation.
No destruction of the lung parenchyma consistent with emphysema.
No evidence of focal or diffuse emphysema.
No picture of an emphysematous lung.
No hypertransparent lung fields consistent with emphysema.
No emphysema-typical increased lung volume.
Lung parenchyma without bullous remodelling.
No apical bullae delineable.
No evidence of emphysema-typical thoracic configuration.
Normal pulmonary vascular markings, no evidence of emphysema.
No signs of destructive parenchymal remodelling.
No radiological criteria for emphysema.
No evidence of emphysema-typical parenchymal rarefaction.
Diaphragm not flattened, no suspicion of emphysema.
Thoracic shape without evidence of emphysematous hyperinflation.
No emphysema bullae or emphysema-typical remodelling.
No signs of emphysema are found.
No emphysema-suspicious finding.

\bigskip
\noindent\textbf{Pathologies -- Lung: Fibrosis}\\
No fibrosis.
No evidence of pulmonary fibrosis.
No fibrotic changes.
No evidence of interstitial changes consistent with fibrosis.
Lung without signs of fibrosis.
No reticular densities.
No evidence of interstitial lung disease.
Lung parenchyma without fibrotic remodelling.
No signs of pulmonary fibrosis bilaterally.
No evidence of fibrotic pulmonary structural changes.
Lung structure unremarkable, no fibrotic changes delineable.
No radiological evidence of fibrotic parenchymal remodelling.
No reticulonodular pattern consistent with fibrosis.
No evidence of subpleural fibrosis.
Lung fields without fibrosis-related increased markings.
No architectural distortion of the lung parenchyma.
No evidence of fibrotic shrinkage changes.
No basal fibrotic remodelling.
No picture of a chronic fibrosing lung disease.
No reticular parenchymal markings of fibrotic aetiology.
Lung without fibrotic residual changes.
No evidence of interstitial fibrosis.
No fibrosis-related volume loss.
No evidence of scarring interstitial changes.
Lung parenchyma without chronic fibrotic remodelling.
No streaky densities consistent with fibrosis.
No scarring parenchymal distortion.
No evidence of fibrotic changes of the lower lung zones.
Lungs bilaterally without interstitial signs of fibrosis.
No fibrotic remodelling of the lung delineable.
No radiological criteria for pulmonary fibrosis.
No fibrotically altered parenchyma detectable.
No linear to reticular densities consistent with fibrosis.
No signs of fibrosis are seen.
No fibrosis-suspicious finding.

\bigskip
\noindent\textbf{Pathologies -- Lung: Mass lesion}\\
No mass lesion detectable.
No evidence of a pulmonary mass.
No nodules delineable.
No evidence of suspicious masses.
Lung without nodular or mass-like densities.
No suspicious focal findings pulmonary.
Lung parenchyma free of mass lesions.
No focal densities consistent with a mass lesion.
No evidence of an intrapulmonary mass.
No tumour-suspicious lesion visible.
No evidence of a pulmonary lesion.
No nodular lesions of the lung.
No suspicious opacity with mass character.
No malignancy-suspicious mass in the thorax.
No evidence of a solid pulmonary lesion.
No peripheral or central nodule identifiable.
No pulmonary mass lesion delineable.
Lung fields without focal opacities.
No delineable intrathoracic pulmonary mass.
No evidence of a suspicious nodular structure.
No mass lesion in the region of the lung fields.
No radiological evidence of a pulmonary tumour.
No solitary pulmonary nodule.
No evidence of a tumour-suspicious focal finding.
Lung parenchyma without suspicious nodular finding.
No nodular or tumorous lesion visible.
No indication of a bronchopulmonary mass.
No pulmonary lesions with suspicion of malignancy.
No mass finding in the imaged lung segments.
Lung without tumour-suspicious lesions or masses.
No suspicious lesion in the region of the lung parenchyma.
No pulmonary nodular or mass finding.
No radiological criteria for a pulmonary mass.
No pulmonary mass lesions are seen.
No mass-suspicious finding pulmonary.

\bigskip
\noindent\textbf{Pathologies -- Lung: Atelectasis}\\
The lung is symmetrically aerated.
The lung is normally aerated.
The lung is adequately aerated.
No ventilatory disturbances detectable.
No areas of reduced aeration detectable.
No atelectasis detectable.
No dyselectasis detectable.
No atelectasis or dyselectasis detectable.
On the radiograph, no ventilatory disturbances are identifiable.
On the radiograph, no areas of reduced aeration are identifiable.
On the radiograph, no atelectasis or dyselectasis is identifiable.
No evidence of atelectasis.
No evidence of dyselectasis.
No evidence of ventilatory disturbances.
No evidence of reduced aeration.
No evidence of ventilatory impairment.
No evidence of reduced aeration.
No radiological evidence of atelectasis.
No atelectatic changes delineable.
No dyselectatic changes delineable.
Lung fields without atelectasis.
Lung fields without areas of reduced aeration.
Lung parenchyma without atelectatic changes.
Lungs bilaterally without ventilatory impairment.
Both lungs are free of atelectasis.
No signs of atelectasis.
No atelectatic parenchymal finding.
No areas of impaired aeration detectable.
No signs of atelectatic areas.

\subsubsection{Internal dataset details}
\label{subsec: appendix_internal_dataset_details}

\paragraph{De-identification and DICOM filtering.}
For the internal TUM UH dataset, patient names and examination dates were removed from text reports, and DICOM metadata was anonymized in accordance with DICOM PS3.15 Annex E. The RIS-imported report data comprised examination specifics, full-text reports (findings and impression), and identification codes (accession number, patient ID). DICOM filtering retained frontal AP/PA and lateral views acquired in erect, semi-erect, and supine positions and was followed by image quality screening.

\paragraph{View-combination distribution.}
Within the included cohort, approximately $56\%$ of examinations contain one frontal and one lateral view, $42\%$ contain only a frontal view, and the remaining cases comprise other scan combinations.

\paragraph{Split statistics and concept-wise text/label coverage.}

The following tables provide further information regarding the internal dataset. Table \ref{tab:dataset_split_overview} contains information about the number of samples/accessions/patients in the used dataset splits (patient-level splitting).
Tables \ref{tab:text_label_split_overview_before_imputation} and \ref{tab:text_label_split_overview_after_65pct_imputation} provide details on the concept-specific text coverage, positive label rate and masked sample percentages in the three dataset splits. 

\begin{table}[htpb]
\centering
\caption{Dataset split overview.}
\label{tab:dataset_split_overview}
\begin{tabular}{lrrr}
\toprule
\textbf{Split} & \textbf{\# Samples} & \textbf{\# Accessions} & \textbf{\# Patients} \\
\midrule
Train & 258{,}236 & 162{,}650 & 91{,}659 \\
Val   & 32{,}766  & 20{,}476  & 11{,}457 \\
Test  & 32{,}560  & 20{,}476  & 11{,}457 \\
\bottomrule
\end{tabular}
\end{table}

\begin{table*}[htpb]
\centering
\caption{Initial text coverage and label statistics by concept and dataset split before any synthetic data imputation. For each split, we report the percentage of samples with text, the positive label rate among valid (unmasked) samples, and the masking rate. If label information is unavailable, entries are marked with ``-''.}
\label{tab:text_label_split_overview_before_imputation}
\scriptsize
\setlength{\tabcolsep}{2pt}
\begin{tabular}{lccc ccc ccc}
\toprule
& \multicolumn{3}{c}{\textbf{Train}} & \multicolumn{3}{c}{\textbf{Val}} & \multicolumn{3}{c}{\textbf{Test}} \\
\cmidrule(lr){2-4} \cmidrule(lr){5-7} \cmidrule(lr){8-10}
\textbf{Field} 
& \shortstack{\textbf{\% Text}} 
& \shortstack{\textbf{Pos. Label}\\\textbf{Rate}} 
& \shortstack{\textbf{\% Label}\\\textbf{Masked}} 
& \shortstack{\textbf{\% Text}} 
& \shortstack{\textbf{Pos. Label}\\\textbf{Rate}} 
& \shortstack{\textbf{\% Label}\\\textbf{Masked}} 
& \shortstack{\textbf{\% Text}} 
& \shortstack{\textbf{Pos. Label}\\\textbf{Rate}} 
& \shortstack{\textbf{\% Label}\\\textbf{Masked}} \\
\midrule

\multicolumn{10}{l}{\textbf{Support Devices}} \\
\midrule
Airway                              & 1.20   & 0.87  & 0.00   & 1.13   & 0.78  & 0.00   & 1.33   & 1.01  & 0.00   \\
Pacemaker                           & 4.70   & 4.45  & 0.00   & 4.89   & 4.63  & 0.00   & 4.49   & 4.26  & 0.00   \\
Feeding Tube                        & 0.61   & 0.56  & 0.00   & 0.68   & 0.62  & 0.00   & 0.74   & 0.66  & 0.00   \\
Other                               & 6.39   & 6.28  & 0.00   & 6.57   & 6.42  & 0.00   & 6.31   & 6.20  & 0.00   \\
Chest Tube                          & 6.71   & 5.90  & 0.00   & 6.31   & 5.52  & 0.00   & 6.52   & 5.76  & 0.00   \\
Central Venous Catheter             & 15.74  & 15.08 & 0.00   & 16.17  & 15.39 & 0.00   & 16.01  & 15.28 & 0.00   \\

\midrule
\multicolumn{10}{l}{\textbf{Pathologies}} \\
\midrule
Pulmonary Edema                     & 0.56   & 62.58 & 99.42  & 0.60   & 63.64 & 99.36  & 0.57   & 53.77 & 99.39  \\
Vascular Congestion                 & 83.64  & 16.34 & 16.34  & 83.67  & 16.34 & 16.30  & 83.60  & 16.27 & 16.37  \\
Fracture                            & 19.63  & 31.58 & 80.34  & 19.12  & 31.18 & 80.84  & 19.16  & 31.31 & 80.79  \\
Other Osseous Findings              & 43.80  & 87.37 & 56.37  & 43.94  & 87.96 & 56.24  & 43.57  & 87.13 & 56.61  \\
Atelectasis                         & 14.88  & 67.46 & 85.07  & 15.07  & 67.98 & 84.85  & 14.55  & 68.26 & 85.41  \\
Emphysema                           & 6.88   & 98.62 & 93.11  & 6.63   & 98.86 & 93.33  & 6.85   & 98.44 & 93.12  \\
Fibrosis                            & 2.58   & 96.96 & 97.40  & 2.90   & 96.03 & 97.08  & 2.41   & 95.37 & 97.55  \\
Pneumonia                           & 91.43  & 17.42 & 8.51   & 91.66  & 17.67 & 8.28   & 91.50  & 17.44 & 8.42   \\
Mass Lesion                               & 18.85  & 40.24 & 81.14  & 19.44  & 40.76 & 80.54  & 18.54  & 40.09 & 81.42  \\
Pleural Effusion                    & 94.07  & 26.29 & 5.89   & 94.25  & 26.33 & 5.69   & 94.13  & 26.08 & 5.82   \\
Pneumothorax                        & 60.37  & 6.43  & 39.54  & 60.49  & 6.33  & 39.37  & 60.33  & 6.21  & 39.55  \\

\midrule
\multicolumn{10}{l}{\textbf{Thoracic Organs}} \\
\midrule
Heart                               & 82.83  & 32.48 & 17.20  & 82.26  & 33.41 & 17.75  & 82.51  & 32.31 & 17.55  \\
Mediastinum                         & 75.55  & 36.20 & 24.38  & 75.51  & 36.11 & 24.43  & 75.45  & 36.21 & 24.49  \\

\midrule
\multicolumn{10}{l}{\textbf{Other}} \\
\midrule
Other mentioned aspects                              & 74.74  & -     & 100.00 & 74.50  & -     & 100.00 & 74.26  & -     & 100.00 \\
\bottomrule
\end{tabular}
\end{table*}

\begin{table*}[htpb]
\centering
\caption{Text coverage and label statistics by concept and dataset split after synthetic text/label imputation as described in the Methods Section. For each split, we report the percentage of samples with text, the positive label rate among valid (unmasked) samples, and the masking rate. If label information is unavailable, entries are marked with ``-''.}
\label{tab:text_label_split_overview_after_65pct_imputation}
\scriptsize
\setlength{\tabcolsep}{2pt}
\begin{tabular}{lccc ccc ccc}
\toprule
& \multicolumn{3}{c}{\textbf{Train}} & \multicolumn{3}{c}{\textbf{Val}} & \multicolumn{3}{c}{\textbf{Test}} \\
\cmidrule(lr){2-4} \cmidrule(lr){5-7} \cmidrule(lr){8-10}
\textbf{Field} 
& \shortstack{\textbf{\% Text}} 
& \shortstack{\textbf{Pos. Label}\\\textbf{Rate}} 
& \shortstack{\textbf{\% Label}\\\textbf{Masked}} 
& \shortstack{\textbf{\% Text}} 
& \shortstack{\textbf{Pos. Label}\\\textbf{Rate}} 
& \shortstack{\textbf{\% Label}\\\textbf{Masked}} 
& \shortstack{\textbf{\% Text}} 
& \shortstack{\textbf{Pos. Label}\\\textbf{Rate}} 
& \shortstack{\textbf{\% Label}\\\textbf{Masked}} \\
\midrule
\multicolumn{10}{l}{\textbf{Support Devices}} \\
\midrule
Airway                              & 2.68   & 0.87  & 0.00   & 2.62   & 0.78  & 0.00   & 2.85   & 1.01  & 0.00   \\
Pacemaker                           & 10.37  & 4.45  & 0.00   & 10.41  & 4.63  & 0.00   & 10.29  & 4.26  & 0.00   \\
Feeding Tube                        & 1.36   & 0.56  & 0.00   & 1.46   & 0.62  & 0.00   & 1.49   & 0.66  & 0.00   \\
Other                               & 14.10  & 6.28  & 0.00   & 14.29  & 6.42  & 0.00   & 14.07  & 6.20  & 0.00   \\
Chest Tube                          & 14.85  & 5.90  & 0.00   & 14.86  & 5.52  & 0.00   & 14.71  & 5.76  & 0.00   \\
Central Venous Catheter             & 35.01  & 15.08 & 0.00   & 35.67  & 15.39 & 0.00   & 35.34  & 15.28 & 0.00   \\
\midrule
\multicolumn{10}{l}{\textbf{Pathologies}} \\
\midrule
Pulmonary Edema                     & 1.33   & 27.10 & 98.67  & 1.41   & 29.89 & 98.64  & 1.35   & 24.49 & 98.66  \\
Vascular Congestion                 & 83.64  & 16.34 & 16.34  & 83.67  & 16.34 & 16.30  & 83.60  & 16.27 & 16.37  \\
Fracture                            & 20.60  & 30.08 & 79.35  & 20.22  & 29.49 & 79.75  & 20.15  & 29.76 & 79.79  \\
Other Osseous Findings              & 43.80  & 87.37 & 56.37  & 43.94  & 87.96 & 56.24  & 43.57  & 87.13 & 56.61  \\
Atelectasis                         & 16.21  & 61.97 & 83.74  & 16.35  & 62.75 & 83.59  & 15.85  & 62.69 & 84.11  \\
Emphysema                           & 8.45   & 80.23 & 91.53  & 8.40   & 78.03 & 91.56  & 8.33   & 81.01 & 91.64  \\
Fibrosis                            & 4.85   & 51.61 & 95.12  & 5.23   & 53.24 & 94.73  & 4.62   & 50.13 & 95.33  \\
Pneumonia                           & 91.43  & 17.42 & 8.51   & 91.66  & 17.67 & 8.28   & 91.50  & 17.44 & 8.42   \\
Mass Lesion                         & 19.74  & 38.42 & 80.25  & 20.25  & 39.16 & 79.74  & 19.27  & 38.58 & 80.69  \\
Pleural Effusion                    & 94.07  & 26.29 & 5.89   & 94.25  & 26.33 & 5.69   & 94.13  & 26.08 & 5.82   \\
Pneumothorax                        & 60.37  & 6.43  & 39.54  & 60.49  & 6.33  & 39.37  & 60.33  & 6.21  & 39.55  \\
\midrule
\multicolumn{10}{l}{\textbf{Thoracic Organs}} \\
\midrule
Heart                               & 82.83  & 32.48 & 17.20  & 82.26  & 33.41 & 17.75  & 82.51  & 32.31 & 17.55  \\
Mediastinum                         & 75.55  & 36.20 & 24.38  & 75.51  & 36.11 & 24.43  & 75.45  & 36.21 & 24.49  \\
\midrule
\multicolumn{10}{l}{\textbf{Other}} \\
\midrule
Other mentioned aspects             & 74.74  & -     & 100.00 & 74.50  & -     & 100.00 & 74.26  & -     & 100.00 \\
\bottomrule
\end{tabular}
\end{table*}

\newpage
\subsubsection{Image augmentation details}
\label{subsubsec: appendix_img_augmentations}

In the training pipeline, image augmentation was restricted to the training set, whereas validation images were processed deterministically without stochastic perturbations. 
The augmentation sequence consisted of a conservative random affine transformation followed by a custom corner cutout and subsequent normalization. Specifically, each chest radiograph was subjected to a random in-plane rotation of up to $ \pm $ 5° and a random translation of up to 2 \% of the image extent in both horizontal and vertical directions; no additional scaling or shearing was applied. Empty regions introduced by the affine transform were filled with zero intensity. After the affine step, images were resized to 518 × 518 pixels. To further reduce potential shortcut learning from peripheral acquisition markers, a custom corner-cutout operation was applied that independently masked rectangular patches covering 20 \% of the image height and 25 \% of the image width in predefined corners. The upper-left corner was masked with probability 0.70 and the upper-right corner with probability 0.85, whereas no masking was applied to the lower corners (probability 0.0 each). Masked regions were filled with zero intensity prior to normalization.

\subsubsection{Label-supervised classification training details}
\label{app:classification-training}

For linear probing concept classification (cf.\ the main-text Methods, \nameref{sec:evaluation}), only the concept-wise linear classification heads were fine-tuned, each taking its corresponding concept-specific visual feature vector as input, while the image encoder and all other alignment-related parameters were frozen. The heads were optimized with a masked binary cross-entropy loss evaluated only on samples with valid LLM-derived labels (i.e., excluding the no-information class). Concept-wise positive class weights derived from the training-label distribution were used to address class imbalance, additionally tempered with $\gamma=0.7$ and combined with concept-balanced reduction to prevent concepts with higher label frequency from dominating the objective.

Training was performed for 14 epochs using AdamW (learning rate $5\cdot 10^{-4}$, weight decay $10^{-4}$) with bfloat16 mixed precision, gradient clipping at $1.0$, and a linear warm-up over the first $1{,}000$ steps followed by cosine learning-rate decay. The best checkpoint was selected based on the validation macro AUPRC.

\newpage
\subsubsection{Zero-shot classification prompt database}

\begin{table*}[htpb]
\centering
\caption{Zero-shot prompting sentences (Translated from German to English) used per concept. For each field, three positive and three negative template sentences are provided as classification anchors.}
\label{tab: appendix_zero_shot_prompts}
\scriptsize
\setlength{\tabcolsep}{3pt}
\renewcommand{\arraystretch}{1.15}
\begin{tabular}{p{0.21\linewidth} p{0.36\linewidth} p{0.36\linewidth}}
\toprule
\textbf{Field / subfield} & \textbf{Positive prompts} & \textbf{Negative prompts} \\
\midrule
\multicolumn{3}{l}{\textbf{Support Devices}} \\
\midrule
Airway
  & (1)~Endotracheal tube present. (2)~Tracheal cannula present. (3)~Airway foreign material is visible.
  & (1)~No endotracheal tube. (2)~No tracheal cannula. (3)~No airway foreign material visible. \\[4pt]
Feeding tube
  & (1)~Gastric tube present. (2)~Enteral tube is visible. (3)~Feeding tube visible in projection.
  & (1)~No gastric tube. (2)~No enteral tube. (3)~No evidence of feeding tube. \\[4pt]
Central venous access
  & (1)~Central venous access present. (2)~CVC present. (3)~Port catheter or central access visible.
  & (1)~No central venous access. (2)~No CVC. (3)~No port catheter visible. \\[4pt]
Chest drain
  & (1)~Chest drain present. (2)~Pleural drain in place. (3)~Drainage tube visible in the thorax.
  & (1)~No chest drain. (2)~No pleural drain. (3)~No drainage tube visible in the thorax. \\[4pt]
Pacemaker
  & (1)~Cardiac pacemaker present. (2)~Pacemaker leads are visible. (3)~ICD or pacemaker is present.
  & (1)~No cardiac pacemaker. (2)~No pacemaker leads visible. (3)~No ICD or pacemaker present. \\[4pt]
Other
  & (1)~Other therapeutic devices present. (2)~Further foreign materials visible. (3)~Additional medical material is present.
  & (1)~No other therapeutic devices. (2)~No further foreign materials. (3)~No additional medical material visible. \\
\midrule
\multicolumn{3}{l}{\textbf{Thoracic Organs}} \\
\midrule
Heart
  & (1)~Cardiac finding abnormal. (2)~Cardiac abnormality present. (3)~Pathological finding of the heart.
  & (1)~Heart without pathological finding. (2)~No cardiac abnormality. (3)~Unremarkable cardiac finding. \\[4pt]
Mediastinum
  & (1)~Mediastinal finding abnormal. (2)~Mediastinal abnormality present. (3)~Pathological finding in the mediastinum.
  & (1)~Mediastinum without pathological finding. (2)~No mediastinal abnormality. (3)~Unremarkable mediastinal finding. \\
\midrule
\multicolumn{3}{l}{\textbf{Pathologies}} \\
\midrule
Lung: Pneumonia
  & (1)~Pneumonia present. (2)~Infiltrate consistent with pneumonia. (3)~Inflammatory infiltration of the lung.
  & (1)~No pneumonia. (2)~No infiltrate consistent with pneumonia. (3)~No evidence of inflammatory infiltration. \\[4pt]
Lung: Atelectasis
  & (1)~Atelectasis present. (2)~Subsegmental atelectasis visible. (3)~Collapsed area of the lung present.
  & (1)~No atelectasis. (2)~No subsegmental atelectasis. (3)~No collapsed area of the lung. \\[4pt]
Lung: Emphysema
  & (1)~Pulmonary emphysema present. (2)~Emphysematous hyperinflation visible. (3)~Hyperinflation due to emphysema.
  & (1)~No pulmonary emphysema. (2)~No emphysematous hyperinflation. (3)~No hyperinflation due to emphysema. \\[4pt]
Lung: Fibrosis
  & (1)~Pulmonary fibrosis present. (2)~Fibrotic changes visible. (3)~Interstitial fibrosis of the lung.
  & (1)~No pulmonary fibrosis. (2)~No fibrotic changes. (3)~No interstitial fibrosis. \\[4pt]
Lung: Mass lesion
  & (1)~Pulmonary mass present. (2)~Suspicious mass in the lung. (3)~Nodule or tumour-suspicious lesion.
  & (1)~No pulmonary mass. (2)~No suspicious mass in the lung. (3)~No nodule and no tumour-suspicious lesion. \\[4pt]
Pleura: Pneumothorax
  & (1)~Pneumothorax present. (2)~Air in the pleural space visible. (3)~Pleural line with pneumothorax.
  & (1)~No pneumothorax. (2)~No air in the pleural space. (3)~No pleural line of a pneumothorax. \\[4pt]
Pleura: Pleural effusion
  & (1)~Pleural effusion present. (2)~Fluid in the pleural space visible. (3)~Basal opacity due to effusion.
  & (1)~No pleural effusion. (2)~No fluid in the pleural space. (3)~No basal effusion opacity. \\[4pt]
Vessels: Congestion
  & (1)~Pulmonary congestion present. (2)~Vascular markings increased due to congestion. (3)~Signs of cardiac congestion.
  & (1)~No pulmonary congestion. (2)~No increased vascular markings. (3)~No signs of cardiac congestion. \\[4pt]
Vessels: Pulmonary edema
  & (1)~Pulmonary oedema present. (2)~Interstitial or alveolar edema visible. (3)~Pulmonary oedema with signs of congestion.
  & (1)~No pulmonary oedema. (2)~No interstitial or alveolar edema. (3)~No pulmonary signs of edema. \\[4pt]
Bones: Fracture
  & (1)~Fracture present. (2)~Rib fracture or osseous fracture visible. (3)~Discontinuity of bone.
  & (1)~No fracture. (2)~No rib fracture. (3)~No osseous discontinuity. \\[4pt]
Bones: Other
  & (1)~Other osseous abnormality present. (2)~Pathological finding of the bony thorax. (3)~Bone changes visible.
  & (1)~No other osseous abnormality. (2)~No pathological finding of the bony thorax. (3)~No bone changes visible. \\
\bottomrule
\end{tabular}
\end{table*}

\subsubsection{Visual grounding evaluation details}
\label{app:grounding-details}

For each evaluated case, the target-concept attention map was bilinearly upsampled to the model input resolution of $518\times 518$ pixels and compared with the corresponding ground-truth CheXlocalize segmentation mask after resizing and letterbox padding to a square canvas; attention values were min--max normalized over the full evaluation canvas.

For the auxiliary thresholded-segmentation analysis, target-specific attention thresholds were determined on the CheXlocalize validation split by sweeping normalized attention thresholds from $0.00$ to $1.00$ in steps of $0.01$ and selecting the thresholds maximizing mean Dice and mean IoU, respectively, across valid cases. These optimized thresholds were then transferred to the test split for evaluation. Whereas the pointing-game hit-rate analysis is threshold-independent, the optimized-threshold procedure standardizes the auxiliary overlap-based comparisons.

Uncertainty for the hit rate, Dice, and IoU was estimated using nonparametric case-level bootstrap resampling: for a concept with $n$ valid instances, $1{,}000$ bootstrap samples of size $n$ were drawn with replacement, the metric was recomputed for each resample, and the bootstrap standard deviation or confidence intervals were used as the uncertainty estimate.

\newpage
\subsubsection{CheXlocalize label class-to-concept mappings}
\label{subsec: appendix_chexpert_label_class_concept_mappings}

\begin{table}[htpb]
\centering
\small
\setlength{\tabcolsep}{4.5pt}
\renewcommand{\arraystretch}{1.15}
\caption{Mapping from structured-report concepts to CheXpert/CheXlocalize label classes used for external classification evaluation.}
\label{tab:mapping_classification}
\begin{tabular}{ll}
\toprule
\textbf{CheXpert / CheXlocalize class} & \textbf{Structured-report concept(s)} \\
\midrule
Support Devices & Support Devices -- Airway, Feeding tube, Central venous \\
                & access, Chest drain, Pacemaker, Other \\[4pt]
Enlarged Cardiomediastinum & Thoracic organs -- Mediastinum \\[4pt]
Cardiomegaly & Thoracic organs -- Heart \\[4pt]
Edema & Pathologies -- Vessels: Congestion, Pulmonary edema  \\[4pt]
Pneumonia & Pathologies -- Lung: Pneumonia \\[4pt]
Atelectasis & Pathologies -- Lung: Atelectasis \\[4pt]
Pneumothorax & Pathologies -- Pleura: Pneumothorax \\[4pt]
Pleural Effusion & Pathologies -- Pleura: Pleural effusion \\[4pt]
Fracture & Pathologies -- Bones: Fracture \\[4pt]
Lung Lesion & Pathologies -- Lung: Mass lesion \\
\bottomrule
\end{tabular}
\end{table}

\begin{table}[htpb]
\centering
\small
\setlength{\tabcolsep}{4.5pt}
\renewcommand{\arraystretch}{1.15}
\caption{Strict one-to-one concept mappings used for the visual grounding evaluation on CheXlocalize.}
\label{tab:mapping_grounding}
\begin{tabular}{ll}
\toprule
\textbf{CheXpert / CheXlocalize class} & \textbf{Structured-report concept} \\
\midrule
Cardiomegaly & Thoracic organs -- Heart \\[4pt]
Enlarged Cardiomediastinum & Thoracic organs -- Mediastinum \\[4pt]
Atelectasis & Pathologies -- Lung: Atelectasis \\[4pt]
Pneumothorax & Pathologies -- Pleura: Pneumothorax \\[4pt]
Pleural Effusion & Pathologies -- Pleura: Pleural effusion \\[4pt]
Lung Lesion & Pathologies -- Lung: Mass lesion \\
\bottomrule
\end{tabular}
\end{table}

\subsubsection{Baseline setup implementation details}
\label{app:baselines-details}

The \emph{report-level alignment baseline} retains the architectural components of our setup (RAD-DINO-MAIRA-2 image encoder; Qwen3-Embedding-4B text embedding with output dimension $768$; further details in the main-text Methods, \nameref{subsec: Implementation details}) but embeds the full unstructured report directly and uses a single query token in the cross-attention block, producing a single global visual feature applied to all concepts. It was trained for a maximum of $20$ epochs (with early validation loss stopping). For zero-shot classification, the same concept-specific prompt database and two-way prototype comparison were used as for the main model (main-text Methods, \nameref{sec:evaluation}), with the global visual feature serving as input for all concepts.
The \emph{linear-probe baseline} fits concept-wise linear classification heads directly on the frozen RAD-DINO-MAIRA-2 CLS output token, without any vision-language pretraining, using the same LLM-derived binary labels. The optimization strategy and parameters are identical to those described in Supplementary Section~\ref{app:classification-training}, except for a maximum of $22$ training epochs.
For the \emph{CARZero baseline}~\cite{lai2024carzero}, classification and grounding evaluations on CheXlocalize were performed based on the open-source benchmark implementation\footnote{\url{https://github.com/Roypic/Benchmarkingattention}} associated with the work by Luo et al.\cite{luo2025xbench}. As CARZero operates at an input resolution of $224\times 224$ pixels, we additionally retrained our setup at this lower resolution (all other implementation aspects retained) for the visual grounding comparison to ensure a fair assessment.

\subsubsection{Reader study sampling protocol (vision-language evaluation)}
\label{app:reader-study-vlm-sampling}

Candidate cases were restricted to frontal radiographs, and concept-wise quota sampling was used to ensure a more balanced label distribution within the reader-study sample pools. For the internal dataset, the quota comprised four samples with a positive label for each support device concept and two positive plus two negative samples for each remaining concept; the remaining cases up to $N=100$ were drawn at random. For the external CheXlocalize dataset, the quota comprised four positive `No Finding' samples and two positive plus two negative samples for each remaining target label, with the remainder again drawn at random up to $N=100$.

\subsubsection{Reader study model output derivation and rating scheme}
\label{app:reader-study-vlm-details}

\textbf{Rating scheme.} For each case and concept, readers selected ``correct'' or ``incorrect'' for the binary label prediction, and ``correct,'' ``partially correct,'' or ``incorrect'' for both the attention map and the top-1 retrieved text. If ``correct'' was not selected for the top-1 text match, the top-2 and top-3 best matches were additionally displayed, and readers could indicate whether one, both, or none of them was correct; a free-text field allowed readers to optionally provide a corrected text proposal.
\textbf{Operating thresholds.} Per-concept operating thresholds for binarizing classifier outputs on the reader-study cases were derived exclusively from the validation split using the best linear probing checkpoint. Sigmoid probabilities were collected for each concept on cases with valid concept labels, candidate thresholds were swept from $0.00$ to $1.00$ in increments of $0.001$, and the operating point maximizing Youden's $J$ statistic was selected for each concept.
\textbf{Retrieval database.} A global concept-wise retrieval database was constructed from the training split: for each training accession and concept (deduplication of identical text spans included), the corresponding precomputed text embedding was projected by the model's learned text projection into the shared vision-language embedding space. At evaluation, the visual embedding of the queried concept token was L2-normalized and compared against all normalized database embeddings of the same concept using cosine similarity, and the highest-scoring unique report snippets were returned as the best-matching descriptions.

\subsection{Supplementary results}

\subsubsection{Internal-dataset evaluation: extended analysis}
\label{subsubsec: appendix_internal_eval_extended}

\paragraph{Per-seed stability.}
Each training configuration was repeated with three random seeds to assess stochastic-optimization variability. Median values were used for the discussion and figures in the main text. For the \emph{Qwen3} setup (RadPRISM), zero-shot macro AUROC across seeds ranged from 0.866 to 0.871 (range 0.005), and linear-probing macro AUROC from 0.890 to 0.891 (range 0.001); the corresponding macro AUPRC ranges were 0.672--0.679 and 0.723--0.726. The \emph{medBERT.de} setup  (text-encoder ablation) showed comparable stability (zero-shot AUROC range 0.002, linear-probing range 0.001; zero-shot AUPRC range 0.014, linear-probing range 0.002), and the unstructured-report baseline was similarly stable (zero-shot AUROC range 0.006; zero-shot AUPRC range 0.005). The same holds for the linear probing baseline (AUROC range 0.001;  AUPRC range 0.002). All inter-setup differences reported in the main text exceed the per-setup seed range by at least an order of magnitude, indicating that the observed contrasts are systematic rather than stochastic.

\paragraph{Qwen3 vs. medBERT.de text-encoder comparison.}
\label{app:qwen-medbert}
The \emph{Qwen3} setup (used RadPRISM configuration) consistently outperformed its \emph{medBERT.de} (text-encoder ablation) counterpart in the zero-shot setting (macro AUROC 0.868 vs.\ 0.860; macro AUPRC 0.676 vs.\ 0.658), while both setups converged to nearly identical linear-probing performance (macro AUROC $\approx$0.891; macro AUPRC $\approx$0.725). The zero-shot advantage of Qwen3 is plausibly attributable to its larger parameter count (4B vs.\ approximately 0.1B) and its training on more diverse multilingual data, both of which produce richer text embeddings of the German concept prompts and thus more discriminative concept prototypes in the shared embedding space. The convergence under linear probing suggests that both text encoders guide the cross-attention module toward comparably informative visual feature extraction during alignment training, and that the encoder-specific quality difference manifests primarily in the text-side embeddings used at zero-shot inference rather than in the learned visual representations themselves.

\paragraph{Concept-wise zero-shot to linear-probing improvements.}
The magnitude of improvement from zero-shot to linear-probing classification varied substantially across concepts. For \textit{Pacemaker} ($\Delta$AUROC\,$=$\,0.001), \textit{Congestion} (0.004), and \textit{Pulmonary edema} (0.005), fine-tuning yielded negligible gains, consistent with already saturated zero-shot AUROC values and well-defined radiographic presentations. The largest gains were observed for \textit{Heart} ($\Delta$AUROC\,$=$\,0.049), \textit{Fracture} (0.044), and \textit{Feeding tube} (0.035). For \textit{Heart}, the relatively generic zero-shot prompts (e.g., ``Cardiac finding abnormal'') yield text prototypes that capture a diffuse region of the embedding space rather than a tight cluster aligned with the dominant clinical pattern (cardiomegaly), whereas linear probing learns a more targeted decision boundary. For \textit{Fracture}, the visual heterogeneity of fracture types and locations and the comparatively low LLM labeling accuracy may jointly limit zero-shot discrimination, while supervised training aggregates the diverse cues despite some label noise. For \textit{Feeding tube}, the very low training prevalence ($\sim$0.6\%) limits exposure during contrastive alignment, which the supervised stage partially compensates for via class-imbalance-aware weighting; co-occurring foreign objects (ECG electrodes, central venous catheters, pacemaker components) in the same image regions may further obscure feeding-tube-specific visual features at the contrastive stage.

\paragraph{AUPRC vs. AUROC pattern.}
Macro AUPRC values were uniformly lower than the corresponding AUROC values across all setups, reflecting the substantial class imbalance present in the dataset. The relative improvement pattern was preserved (concept-stratified vs.\ unstructured-report baseline: zero-shot AUPRC 0.676 vs.\ 0.451; concept-stratified vs. linear-probing baseline 0.725 vs.\ 0.642). The larger absolute AUROC--AUPRC gap for the global-feature baseline (0.717 vs.\ 0.451) compared to the concept-stratified setup (0.868 vs.\ 0.676) suggests that the global-feature model is particularly challenged by low-prevalence concepts, where precision at clinically relevant recall thresholds is critical.

\newpage
\subsubsection{Quantitative visual grounding evaluation}

\begin{table}[!ht]
\centering
\small
\setlength{\tabcolsep}{4.5pt}
\renewcommand{\arraystretch}{1.15}
\caption{Visual grounding performance of \emph{RadPRISM} (518 px img res) on the CheXlocalize test split evaluated via Dice score and IoU (Intersection over Union) (1\,000 bootstrap samples used for the std. dev. estimation). Thresholds used on the test split were optimized on the validation split.}
\label{tab: appendix_grounding_dice_iou}
\begin{tabular}{l cc cc}
\toprule
& \multicolumn{2}{c}{\textbf{Dice}} & \multicolumn{2}{c}{\textbf{IoU}} \\
\cmidrule(lr){2-3} \cmidrule(lr){4-5}
\textbf{CheXlocalize class} & \textbf{Mean} & \textbf{Standard Deviation} & \textbf{Mean} & \textbf{Standard Deviation} \\
\midrule
Cardiomegaly              & 0.614 & 0.010 & 0.455 & 0.010 \\[3pt]
Enlarged Cardiomediastinum & 0.509 & 0.007 & 0.351 & 0.006 \\[3pt]
Lung Lesion               & 0.398 & 0.073 & 0.290 & 0.064 \\[3pt]
Atelectasis               & 0.394 & 0.014 & 0.262 & 0.011 \\[3pt]
Pleural Effusion          & 0.353 & 0.017 & 0.232 & 0.014 \\[3pt]
Pneumothorax              & 0.131 & 0.025 & 0.034 & 0.010 \\
\bottomrule
\end{tabular}
\end{table}

\newpage
\subsubsection{Additional qualitative examples}

\begin{figure}[htbp]
  \centering
  \includegraphics[width=\linewidth]{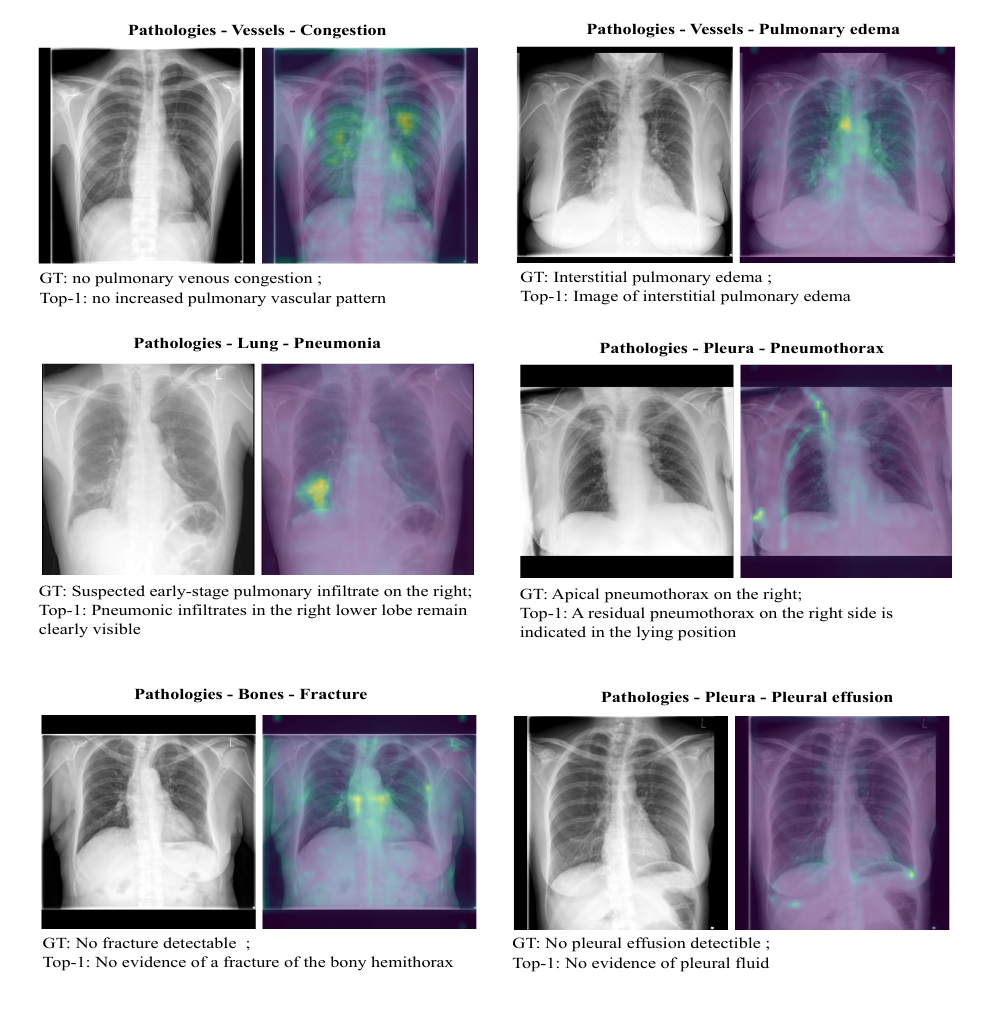}
  \caption{Extended examples for the concept-specific attention map grounding and the top-1 retrieval with comparisons to the ground truth (GT).}
  \label{fig: appendix_further_attn_map_examples1}
\end{figure}

\begin{figure}[htbp]
  \centering
  \includegraphics[width=\linewidth]{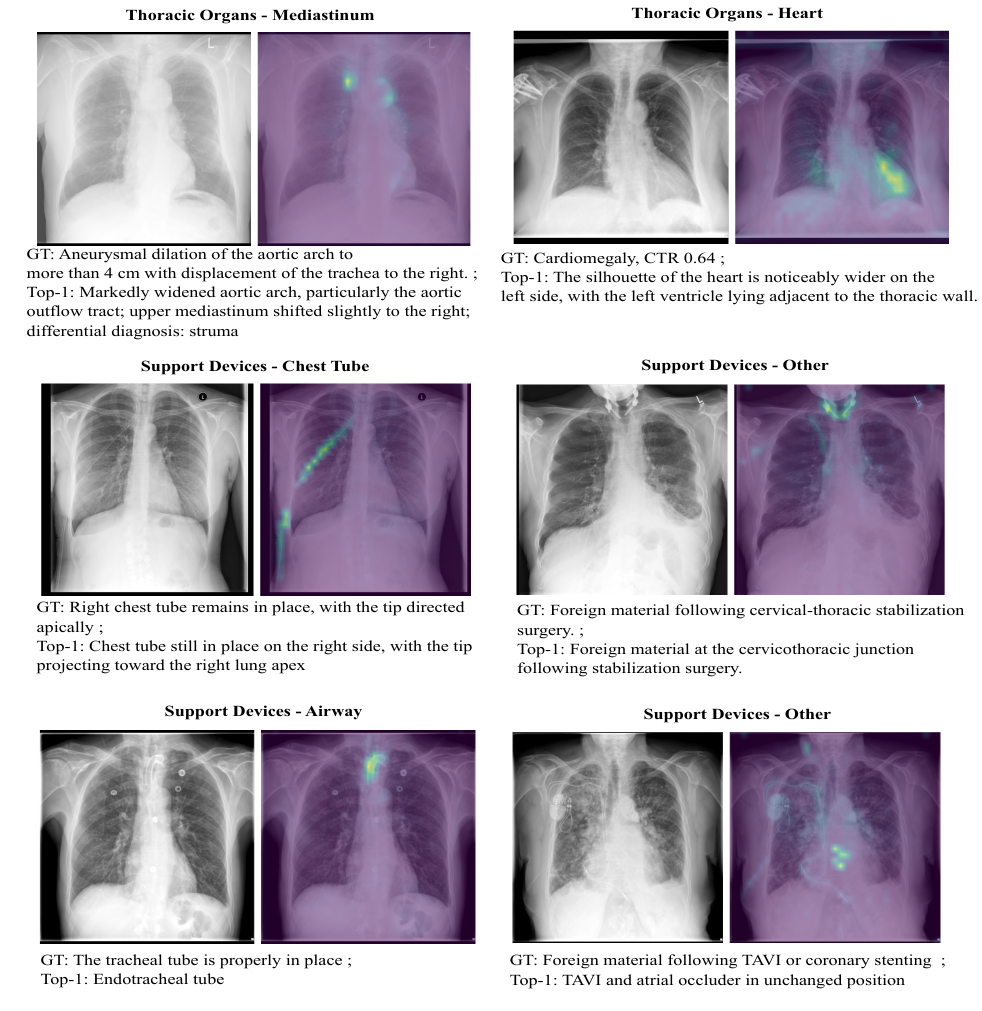}
  \caption{Extended examples for the concept-specific attention map grounding and the top-1 retrieval with comparisons to the ground truth (GT).}
  \label{fig: appendix_further_attn_map_examples2}
\end{figure}

\newpage

\subsubsection{Structuring and labeling reader study}

\begin{table}[htpb]
  \centering
  \small
  \setlength{\tabcolsep}{3.5pt}
  \renewcommand{\arraystretch}{1.15}
  \caption{Reader study evaluation for structured report fields (per-reader). Metrics for labels: macro-$F_1$, Cohen's $\kappa$, and metric for text entries: BERTScore $F_1$.}
  \label{tab:reader-study-metrics}
  \begin{tabular}{p{0.33\linewidth} ccc ccc ccc}
    \toprule
    & \multicolumn{3}{c}{macro $F_1$} & \multicolumn{3}{c}{$\kappa$} & \multicolumn{3}{c}{BERTScore $F_1$} \\
    \cmidrule(lr){2-4}\cmidrule(lr){5-7}\cmidrule(lr){8-10}
    Field / subfield & R1 & R2 & R3 & R1 & R2 & R3 & R1 & R2 & R3 \\
    \midrule
    Examination & -- & -- & -- & -- & -- & -- & 1.00 & 1.00 & 1.00 \\
    Clinical information & -- & -- & -- & -- & -- & -- & 0.81 & 0.91 & 0.92 \\
    Report date & -- & -- & -- & -- & -- & -- & -- & -- & -- \\
    Comparison & 0.99 & 0.99 & 0.99 & 0.99 & 0.98 & 0.98 & 0.94 & 0.97 & 0.95 \\
    \textbf{Support devices} & & & & & & & & & \\
    \quad Airway & 0.97 & 0.97 & 0.97 & 0.93 & 0.93 & 0.93 & 0.94 & 0.93 & 0.91 \\
    \quad Feeding tube & 0.98 & 0.98 & 0.98 & 0.95 & 0.95 & 0.95 & 1.00 & 1.00 & 0.95 \\
    \quad Central venous access & 0.98 & 0.97 & 0.97 & 0.97 & 0.93 & 0.93 & 0.95 & 0.86 & 0.87 \\
    \quad Chest drain & 0.96 & 0.96 & 0.96 & 0.91 & 0.91 & 0.91 & 0.96 & 0.78 & 0.71 \\
    \quad Pacemaker & 0.98 & 0.98 & 0.98 & 0.97 & 0.97 & 0.97 & 0.86 & 0.72 & 0.72 \\
    \quad Other & 0.84 & 0.80 & 0.82 & 0.69 & 0.61 & 0.64 & 0.45 & 0.44 & 0.37 \\
    Heart & 0.78 & 0.84 & 0.89 & 0.77 & 0.82 & 0.89 & 0.88 & 0.87 & 0.88 \\
    Mediastinum & 0.67 & 0.67 & 0.79 & 0.67 & 0.68 & 0.80 & 0.70 & 0.74 & 0.77 \\
    \textbf{Pathologies} & & & & & & & & & \\
    \quad Lung: Pneumonia & 0.90 & 0.85 & 0.95 & 0.91 & 0.87 & 0.94 & 0.90 & 0.93 & 0.92 \\
    \quad Lung: Atelectasis & 0.86 & 0.77 & 0.88 & 0.86 & 0.69 & 0.87 & 0.78 & 0.58 & 0.73 \\
    \quad Lung: Emphysema & 0.91 & 0.89 & 0.95 & 0.93 & 0.93 & 0.96 & 0.91 & 0.94 & 0.92 \\
    \quad Lung: Fibrosis & 0.80 & 0.73 & 0.94 & 0.81 & 0.72 & 0.93 & 0.68 & 0.62 & 0.82 \\
    \quad Lung: Mass lesion & 0.87 & 0.78 & 0.92 & 0.87 & 0.83 & 0.92 & 0.76 & 0.76 & 0.77 \\
    \quad Pleura: Pneumothorax & 0.91 & 0.86 & 0.96 & 0.89 & 0.85 & 0.87 & 0.88 & 0.88 & 0.86 \\
    \quad Pleura: Pleural effusion & 0.87 & 0.85 & 0.91 & 0.89 & 0.86 & 0.91 & 0.87 & 0.95 & 0.84 \\
    \quad Vessels: Congestion & 0.92 & 0.90 & 0.96 & 0.90 & 0.89 & 0.93 & 0.89 & 0.94 & 0.88 \\
    \quad Vessels: Pulmonary edema & 0.83 & 0.71 & 0.83 & 0.67 & 0.59 & 0.77 & 0.48 & 0.44 & 0.62 \\
    \quad Bones: Fracture & 0.71 & 0.68 & 0.74 & 0.87 & 0.88 & 0.97 & 0.79 & 0.86 & 0.90 \\
    \quad Bones: Other & 0.74 & 0.81 & 0.91 & 0.77 & 0.84 & 0.94 & 0.74 & 0.86 & 0.82 \\
    Other findings & -- & -- & -- & -- & -- & -- & 0.40 & 0.34 & 0.38 \\
    Impression & -- & -- & -- & -- & -- & -- & 1.00 & 0.98 & 0.98 \\
    \bottomrule
  \end{tabular}
\end{table}

\begin{table}[t]
  \centering
  \small
  \setlength{\tabcolsep}{4.5pt}
  \renewcommand{\arraystretch}{1.15}
  \caption{Inter-rater reliability per structured report field (N=200). Krippendorff's $\alpha$ and Gwet's AC1 are computed per label field. Fields without label are marked with --.}
  \label{tab:irr-per-field}
  \begin{tabular}{p{0.46\linewidth} l cc}
    \toprule
    Field / subfield & Label type & Krippendorff's $\alpha$ & Gwet's AC1 \\
    \midrule
    Examination & -- & -- & -- \\
    Clinical information & -- & -- & -- \\
    Report date & -- & -- & -- \\
    Comparison & binary & 0.980 & 0.980 \\
    \textbf{Therapy aids} & & & \\
    \quad Airway & binary & 1.000 & 1.000 \\
    \quad Feeding tube & binary & 1.000 & 1.000 \\
    \quad Central venous access & binary & 0.977 & 0.991 \\
    \quad Chest drain & binary & 1.000 & 1.000 \\
    \quad Pacemaker & binary & 1.000 & 1.000 \\
    \quad Other & binary & 0.673 & 0.911 \\
    Heart & multiclass & 0.860 & 0.688 \\
    Mediastinum & multiclass & 0.621 & -0.063 \\
    \textbf{Pathologies} & & & \\
    \quad Lung: Pneumonia & multiclass & 0.889 & 0.909 \\
    \quad Lung: Atelectasis & multiclass & 0.672 & 0.776 \\
    \quad Lung: Emphysema & multiclass & 0.928 & 0.972 \\
    \quad Lung: Fibrosis & multiclass & 0.798 & 0.974 \\
    \quad Lung: Mass lesion & multiclass & 0.843 & 0.940 \\
    \quad Pleura: Pneumothorax & multiclass & 0.883 & 0.862 \\
    \quad Pleura: Pleural effusion & multiclass & 0.870 & 0.848 \\
    \quad Vessels: Congestion & multiclass & 0.893 & 0.872 \\
    \quad Vessels: Pulmonary edema & multiclass & 0.500 & 0.923 \\
    \quad Bones: Fracture & multiclass & 0.817 & 0.887 \\
    \quad Bones: Other & multiclass & 0.779 & 0.722 \\
    Other findings & -- & -- & -- \\
    Impression & -- & -- & -- \\
    \bottomrule
  \end{tabular}
\end{table}

\begin{table}[htpb]
\centering
\small
\setlength{\tabcolsep}{4.5pt}
\renewcommand{\arraystretch}{1.15}
\caption{Inter-reader agreement measured by pairwise BERTScore F1 on the text fields of the structured reports (N\,=\,3 reader pairs, 200 reports).}
\label{tab: appendix_interreader_bertscore_f1}
\begin{tabular}{p{0.46\linewidth} cc}
\toprule
\textbf{Field / subfield} & \textbf{Mean BERTScore F1} & \textbf{Standard Deviation} \\
\midrule
Examination & 0.992 & 0.003 \\
Impression & 0.988 & 0.008 \\
Airway & 0.979 & 0.014 \\
Feeding tube & 0.964 & 0.031 \\
Comparison & 0.961 & 0.003 \\
\textbf{Thoracic organs} & & \\
\quad Heart & 0.907 & 0.015 \\
\quad Mediastinum & 0.856 & 0.028 \\
Clinical information & 0.898 & 0.048 \\
\textbf{Pathologies} & & \\
\quad Lung: Pneumonia & 0.903 & 0.004 \\
\quad Lung: Emphysema & 0.896 & 0.029 \\
\quad Vessels: Congestion & 0.879 & 0.017 \\
\quad Pleura: Pneumothorax & 0.867 & 0.047 \\
\quad Pleura: Pleural effusion & 0.860 & 0.022 \\
\quad Bones: Other & 0.777 & 0.071 \\
\quad Bones: Fracture & 0.753 & 0.073 \\
\quad Lung: Mass lesion & 0.717 & 0.037 \\
\quad Lung: Fibrosis & 0.680 & 0.080 \\
\quad Lung: Atelectasis & 0.591 & 0.077 \\
\quad Vessels: Pulmonary edema & 0.364 & 0.133 \\
\textbf{Support Devices} & & \\
\quad Central venous access & 0.885 & 0.071 \\
\quad Pacemaker & 0.879 & 0.084 \\
\quad Chest drain & 0.784 & 0.060 \\
\quad Other & 0.559 & 0.043 \\
Other findings & 0.479 & 0.050 \\
\bottomrule
\end{tabular}
\end{table}

\newpage
\subsubsection{Vision-language reader study: extended analysis}

\begin{table}[htpb]
  \centering
  \footnotesize
  \setlength{\tabcolsep}{4.5pt}
  \renewcommand{\arraystretch}{1.12}
  \caption{Per-concept \textbf{classification feedback}. For each dataset, readers rated the binarized concept-label prediction as correct or incorrect. $N$ is the number of rated cases; Correct, Incorrect, and No feedback (including feedback missing or not possible) are given as percentages of $N$ (row-wise sum $=100\%$). Rows are pooled across all readers.}
  \label{tab:appendix_vlm_classification_feedback}
  \begin{tabular}{p{0.30\linewidth} rrrr rrrr}
    \toprule
    & \multicolumn{4}{c}{CheXpert} & \multicolumn{4}{c}{Internal} \\
    \cmidrule(lr){2-5}\cmidrule(lr){6-9}
    Concept & $N$ & Corr. & Inc. & No fb. & $N$ & Corr. & Inc. & No fb. \\
    \midrule
    \textbf{Support devices} & & & & & & & & \\
    \quad Airway & 100 & 95.0 & 5.0 & 0.0 & 96 & 91.7 & 7.3 & 1.0 \\
    \quad Feeding tube & 100 & 77.0 & 21.0 & 2.0 & 95 & 72.6 & 26.3 & 1.1 \\
    \quad Central venous access & 100 & 97.0 & 3.0 & 0.0 & 95 & 99.0 & 1.1 & 0.0 \\
    \quad Chest drain & 98 & 81.6 & 15.3 & 3.1 & 94 & 91.5 & 5.3 & 3.2 \\
    \quad Pacemaker & 99 & 98.0 & 2.0 & 0.0 & 95 & 96.8 & 3.2 & 0.0 \\
    \quad Other & 98 & 84.7 & 15.3 & 0.0 & 95 & 88.4 & 10.5 & 1.1 \\
    \textbf{Thoracic organs} & & & & & & & & \\
    \quad Heart & 98 & 84.7 & 10.2 & 5.1 & 94 & 85.1 & 10.6 & 4.3 \\
    \quad Mediastinum & 97 & 77.3 & 21.6 & 1.0 & 94 & 79.8 & 17.0 & 3.2 \\
    \textbf{Pathologies} & & & & & & & & \\
    \quad Lung: Pneumonia & 98 & 88.8 & 9.2 & 2.0 & 93 & 83.9 & 11.8 & 4.3 \\
    \quad Lung: Atelectasis & 98 & 93.9 & 5.1 & 1.0 & 93 & 87.1 & 11.8 & 1.1 \\
    \quad Lung: Emphysema & 97 & 69.1 & 29.9 & 1.0 & 93 & 78.5 & 15.0 & 6.5 \\
    \quad Lung: Fibrosis & 97 & 72.2 & 26.8 & 1.0 & 90 & 75.6 & 15.6 & 8.9 \\
    \quad Lung: Mass lesion & 98 & 75.5 & 21.4 & 3.1 & 92 & 66.3 & 28.3 & 5.4 \\
    \quad Pleura: Pneumothorax & 97 & 80.4 & 18.6 & 1.0 & 92 & 80.4 & 17.4 & 2.2 \\
    \quad Pleura: Pleural effusion & 98 & 89.8 & 9.2 & 1.0 & 93 & 87.1 & 9.7 & 3.2 \\
    \quad Vessels: Congestion & 98 & 85.7 & 13.3 & 1.0 & 93 & 93.5 & 6.5 & 0.0 \\
    \quad Vessels: Pulmonary edema & 98 & 55.1 & 43.9 & 1.0 & 93 & 72.0 & 23.7 & 4.3 \\
    \quad Bones: Fracture & 98 & 59.2 & 39.8 & 1.0 & 93 & 47.3 & 48.4 & 4.3 \\
    \quad Bones: Other & 95 & 73.7 & 23.2 & 3.2 & 92 & 64.1 & 29.3 & 6.5 \\
    \midrule
    \textit{Macro average} & 1862 & 81.0 & 17.6 & 1.5 & 1775 & 81.1 & 15.7 & 3.2 \\
    \bottomrule
  \end{tabular}
\end{table}

\begin{table}[htpb]
  \centering
  \footnotesize
  \setlength{\tabcolsep}{3.5pt}
  \renewcommand{\arraystretch}{1.12}
  \caption{Per-concept \textbf{attention-map feedback}. Readers rated the concept-specific attention map as correct, partially correct, or incorrect. $N$ is the number of rated cases; Correct (Corr.), Partial (Part.), Incorrect (Inc.), and No feedback (No fb.; missing or not possible) are percentages of $N$ (row-wise sum $=100\%$). Metrics are pooled across all readers.}
  \label{tab:appendix_vlm_attention_feedback}
  \begin{tabular}{p{0.26\linewidth} rrrrr rrrrr}
    \toprule
    & \multicolumn{5}{c}{CheXpert} & \multicolumn{5}{c}{Internal} \\
    \cmidrule(lr){2-6}\cmidrule(lr){7-11}
    Concept & $N$ & Corr. & Part. & Inc. & No fb. & $N$ & Corr. & Part. & Inc. & No fb. \\
    \midrule
    \textbf{Support devices} & & & & & & & & & & \\
    \quad Airway & 100 & 93.0 & 6.0 & 1.0 & 0.0 & 96 & 91.7 & 5.2 & 1.0 & 2.1 \\
    \quad Feeding tube & 100 & 73.0 & 25.0 & 0.0 & 2.0 & 95 & 69.5 & 26.3 & 3.2 & 1.1 \\
    \quad Central venous access & 100 & 91.0 & 9.0 & 0.0 & 0.0 & 95 & 90.5 & 9.5 & 0.0 & 0.0 \\
    \quad Chest drain & 98 & 76.5 & 17.3 & 4.1 & 2.0 & 94 & 81.9 & 16.0 & 0.0 & 2.1 \\
    \quad Pacemaker & 99 & 75.8 & 18.2 & 6.1 & 0.0 & 95 & 83.2 & 14.7 & 2.1 & 0.0 \\
    \quad Other & 98 & 88.8 & 10.2 & 1.0 & 0.0 & 95 & 84.2 & 12.6 & 2.1 & 1.1 \\
    \textbf{Thoracic organs} & & & & & & & & & & \\
    \quad Heart & 98 & 92.9 & 1.0 & 1.0 & 5.1 & 94 & 91.5 & 4.3 & 1.1 & 3.2 \\
    \quad Mediastinum & 97 & 87.6 & 8.2 & 3.1 & 1.0 & 94 & 86.2 & 10.6 & 0.0 & 3.2 \\
    \textbf{Pathologies} & & & & & & & & & & \\
    \quad Lung: Pneumonia & 98 & 90.8 & 8.2 & 1.0 & 0.0 & 93 & 93.5 & 1.1 & 0.0 & 5.4 \\
    \quad Lung: Atelectasis & 98 & 86.7 & 9.2 & 3.1 & 1.0 & 93 & 90.3 & 8.6 & 0.0 & 1.1 \\
    \quad Lung: Emphysema & 97 & 77.3 & 19.6 & 2.1 & 1.0 & 93 & 86.0 & 7.5 & 1.1 & 5.4 \\
    \quad Lung: Fibrosis & 97 & 88.7 & 8.2 & 2.1 & 1.0 & 90 & 86.7 & 4.4 & 0.0 & 8.9 \\
    \quad Lung: Mass lesion & 98 & 78.6 & 14.3 & 4.1 & 3.1 & 92 & 79.3 & 13.0 & 2.2 & 5.4 \\
    \quad Pleura: Pneumothorax & 97 & 54.6 & 21.6 & 22.7 & 1.0 & 92 & 66.3 & 15.2 & 16.3 & 2.2 \\
    \quad Pleura: Pleural effusion & 98 & 90.8 & 8.2 & 1.0 & 0.0 & 93 & 93.5 & 3.2 & 0.0 & 3.2 \\
    \quad Vessels: Congestion & 98 & 99.0 & 0.0 & 1.0 & 0.0 & 93 & 98.9 & 1.1 & 0.0 & 0.0 \\
    \quad Vessels: Pulmonary edema & 98 & 75.5 & 20.4 & 3.1 & 1.0 & 93 & 89.2 & 8.6 & 1.1 & 1.1 \\
    \quad Bones: Fracture & 98 & 70.4 & 17.3 & 10.2 & 2.0 & 93 & 68.8 & 16.1 & 11.8 & 3.2 \\
    \quad Bones: Other & 95 & 69.5 & 17.9 & 9.5 & 3.2 & 92 & 68.5 & 22.8 & 2.2 & 6.5 \\
    \midrule
    \textit{Macro average} & 1862 & 82.1 & 12.6 & 4.0 & 1.2 & 1775 & 84.2 & 10.6 & 2.3 & 2.9 \\
    \bottomrule
  \end{tabular}
\end{table}

\begin{table}[htpb]
  \centering
  \scriptsize
  \setlength{\tabcolsep}{2.6pt}
  \renewcommand{\arraystretch}{1.1}
  \caption{Per-concept \textbf{retrieval feedback}. \emph{Top-1 rating}: readers rated the top-1 retrieved report snippet as correct (Corr.), partially correct (Part.), or incorrect (Inc.) over $N$ rated cases (the small No-feedback remainder, $\le$3\%, is omitted here so rows need not sum to $100\%$). \emph{Alternative review}: for the $N_{a}$ cases whose top-1 was not correct, the top-2 and top-3 candidates were additionally shown; Rank 2 / Rank 3 give the percentage of $N_{a}$ for which that candidate was marked correct (readers could mark both), and None the percentage for which no alternative was correct (remainder = no selection). Pooled across all readers.}
  \label{tab:appendix_vlm_retrieval_feedback}
  \begin{tabular}{p{0.185\linewidth} rrrr rrrr rrrr rrrr}
    \toprule
    & \multicolumn{8}{c}{CheXpert} & \multicolumn{8}{c}{Internal} \\
    \cmidrule(lr){2-9}\cmidrule(lr){10-17}
    & \multicolumn{4}{c}{Top-1 rating} & \multicolumn{4}{c}{Alternative review} & \multicolumn{4}{c}{Top-1 rating} & \multicolumn{4}{c}{Alternative review} \\
    \cmidrule(lr){2-5}\cmidrule(lr){6-9}\cmidrule(lr){10-13}\cmidrule(lr){14-17}
    Concept & $N$ & Corr. & Part. & Inc. & $N_{a}$ & R2 & R3 & None & $N$ & Corr. & Part. & Inc. & $N_{a}$ & R2 & R3 & None \\
    \midrule
    \textbf{Support devices} & & & & & & & & & & & & & & & & \\
    \quad Airway & 100 & 81.0 & 11.0 & 8.0 & 19 & 21.1 & 0.0 & 73.7 & 96 & 81.2 & 11.5 & 6.2 & 17 & 35.3 & 35.3 & 29.4 \\
    \quad Feeding tube & 100 & 64.0 & 5.0 & 29.0 & 34 & 8.8 & 8.8 & 58.8 & 95 & 68.4 & 4.2 & 26.3 & 29 & 6.9 & 6.9 & 65.5 \\
    \quad Central venous access & 100 & 83.0 & 15.0 & 2.0 & 17 & 5.9 & 29.4 & 29.4 & 95 & 88.4 & 9.5 & 2.1 & 11 & 36.4 & 27.3 & 45.5 \\
    \quad Chest drain & 98 & 53.1 & 13.3 & 31.6 & 44 & 11.4 & 22.7 & 56.8 & 94 & 81.9 & 8.5 & 6.4 & 14 & 0.0 & 7.1 & 78.6 \\
    \quad Pacemaker & 99 & 88.9 & 7.1 & 4.0 & 11 & 54.5 & 36.4 & 27.3 & 95 & 93.7 & 6.3 & 0.0 & 6 & 33.3 & 16.7 & 66.7 \\
    \quad Other & 98 & 44.9 & 27.6 & 27.6 & 54 & 22.2 & 25.9 & 44.4 & 95 & 51.6 & 12.6 & 34.7 & 45 & 13.3 & 11.1 & 60.0 \\
    \textbf{Thoracic organs} & & & & & & & & & & & & & & & & \\
    \quad Heart & 98 & 62.2 & 18.4 & 15.3 & 33 & 15.2 & 15.2 & 39.4 & 94 & 64.9 & 20.2 & 10.6 & 29 & 31.0 & 48.3 & 20.7 \\
    \quad Mediastinum & 97 & 63.9 & 19.6 & 15.5 & 34 & 14.7 & 23.5 & 35.3 & 94 & 63.8 & 17.0 & 17.0 & 32 & 18.8 & 15.6 & 56.2 \\
    \textbf{Pathologies} & & & & & & & & & & & & & & & & \\
    \quad Lung: Pneumonia & 98 & 64.3 & 24.5 & 11.2 & 35 & 42.9 & 20.0 & 28.6 & 93 & 75.3 & 11.8 & 8.6 & 19 & 26.3 & 21.1 & 42.1 \\
    \quad Lung: Atelectasis & 98 & 69.4 & 22.4 & 7.1 & 29 & 37.9 & 13.8 & 37.9 & 93 & 66.7 & 23.7 & 8.6 & 30 & 26.7 & 40.0 & 30.0 \\
    \quad Lung: Emphysema & 97 & 54.6 & 13.4 & 30.9 & 43 & 11.6 & 18.6 & 58.1 & 93 & 57.0 & 8.6 & 28.0 & 34 & 29.4 & 17.6 & 47.1 \\
    \quad Lung: Fibrosis & 97 & 58.8 & 17.5 & 22.7 & 39 & 23.1 & 20.5 & 46.2 & 90 & 56.7 & 17.8 & 16.7 & 31 & 16.1 & 19.4 & 38.7 \\
    \quad Lung: Mass lesion & 98 & 55.1 & 13.3 & 28.6 & 41 & 19.5 & 26.8 & 41.5 & 92 & 51.1 & 13.0 & 30.4 & 40 & 17.5 & 10.0 & 60.0 \\
    \quad Pleura: Pneumothorax & 97 & 59.8 & 16.5 & 22.7 & 38 & 26.3 & 18.4 & 44.7 & 92 & 65.2 & 17.4 & 15.2 & 30 & 30.0 & 23.3 & 43.3 \\
    \quad Pleura: Pleural effusion & 98 & 68.4 & 16.3 & 15.3 & 31 & 16.1 & 16.1 & 48.4 & 93 & 69.9 & 14.0 & 12.9 & 25 & 16.0 & 20.0 & 40.0 \\
    \quad Vessels: Congestion & 98 & 71.4 & 19.4 & 9.2 & 28 & 42.9 & 39.3 & 28.6 & 93 & 81.7 & 15.0 & 3.2 & 17 & 17.6 & 29.4 & 47.1 \\
    \quad Vessels: Pulmonary edema & 98 & 56.1 & 11.2 & 32.6 & 43 & 11.6 & 9.3 & 58.1 & 93 & 69.9 & 9.7 & 19.4 & 27 & 7.4 & 14.8 & 51.8 \\
    \quad Bones: Fracture & 98 & 64.3 & 2.0 & 31.6 & 33 & 15.2 & 12.1 & 60.6 & 93 & 48.4 & 8.6 & 39.8 & 45 & 8.9 & 8.9 & 53.3 \\
    \quad Bones: Other & 95 & 59.0 & 6.3 & 31.6 & 36 & 13.9 & 16.7 & 47.2 & 92 & 46.7 & 17.4 & 30.4 & 44 & 11.4 & 4.5 & 47.7 \\
    \midrule
    \textit{Macro average} & 1862 & 64.3 & 14.7 & 19.8 & 642 & 21.8 & 19.7 & 45.5 & 1775 & 67.5 & 13.0 & 16.7 & 525 & 20.1 & 19.9 & 48.6 \\
    \bottomrule
  \end{tabular}
\end{table}

\begin{table}[htpb]
  \centering
  \footnotesize
  \setlength{\tabcolsep}{5pt}
  \renewcommand{\arraystretch}{1.12}
  \caption{Per-concept \textbf{retrieval@$k$}. Retrieval@$k$ is the percentage of rated cases for which at least one correct report snippet appeared within the top-$k$ retrieved candidates ($k\in\{1,2,3\}$; values are cumulative and non-decreasing in $k$). $N$ is the number of answered cases used as the denominator. Metrics are pooled across all readers.}
  \label{tab:appendix_vlm_retrieval_at_k}
  \begin{tabular}{p{0.30\linewidth} rrrr rrrr}
    \toprule
    & \multicolumn{4}{c}{CheXpert} & \multicolumn{4}{c}{Internal} \\
    \cmidrule(lr){2-5}\cmidrule(lr){6-9}
    Concept & $N$ & R@1 & R@2 & R@3 & $N$ & R@1 & R@2 & R@3 \\
    \midrule
    \textbf{Support devices} & & & & & & & & \\
    \quad Airway & 100 & 81.0 & 85.0 & 85.0 & 95 & 82.1 & 88.4 & 92.6 \\
    \quad Feeding tube & 98 & 65.3 & 68.4 & 71.4 & 94 & 69.2 & 71.3 & 73.4 \\
    \quad Central venous access & 100 & 83.0 & 84.0 & 88.0 & 95 & 88.4 & 92.6 & 94.7 \\
    \quad Chest drain & 96 & 54.2 & 59.4 & 64.6 & 91 & 84.6 & 84.6 & 85.7 \\
    \quad Pacemaker & 99 & 88.9 & 95.0 & 96.0 & 95 & 93.7 & 95.8 & 95.8 \\
    \quad Other & 98 & 44.9 & 57.1 & 67.3 & 94 & 52.1 & 58.5 & 60.6 \\
    \textbf{Thoracic organs} & & & & & & & & \\
    \quad Heart & 94 & 64.9 & 70.2 & 75.5 & 90 & 67.8 & 77.8 & 90.0 \\
    \quad Mediastinum & 96 & 64.6 & 69.8 & 78.1 & 92 & 65.2 & 71.7 & 75.0 \\
    \textbf{Pathologies} & & & & & & & & \\
    \quad Lung: Pneumonia & 98 & 64.3 & 79.6 & 82.7 & 89 & 78.6 & 84.3 & 87.6 \\
    \quad Lung: Atelectasis & 97 & 70.1 & 81.4 & 83.5 & 92 & 67.4 & 76.1 & 85.9 \\
    \quad Lung: Emphysema & 96 & 55.2 & 60.4 & 67.7 & 87 & 60.9 & 72.4 & 75.9 \\
    \quad Lung: Fibrosis & 96 & 59.4 & 68.8 & 74.0 & 82 & 62.2 & 68.3 & 75.6 \\
    \quad Lung: Mass lesion & 95 & 56.8 & 65.3 & 76.8 & 87 & 54.0 & 62.1 & 64.4 \\
    \quad Pleura: Pneumothorax & 96 & 60.4 & 70.8 & 75.0 & 90 & 66.7 & 76.7 & 80.0 \\
    \quad Pleura: Pleural effusion & 98 & 68.4 & 73.5 & 77.5 & 90 & 72.2 & 76.7 & 80.0 \\
    \quad Vessels: Congestion & 98 & 71.4 & 83.7 & 89.8 & 93 & 81.7 & 85.0 & 89.2 \\
    \quad Vessels: Pulmonary edema & 98 & 56.1 & 61.2 & 64.3 & 92 & 70.7 & 72.8 & 77.2 \\
    \quad Bones: Fracture & 96 & 65.6 & 70.8 & 72.9 & 90 & 50.0 & 54.4 & 57.8 \\
    \quad Bones: Other & 92 & 60.9 & 66.3 & 70.7 & 87 & 49.4 & 55.2 & 57.5 \\
    \midrule
    \textit{Macro average} & 1841 & 65.0 & 72.1 & 76.9 & 1725 & 69.3 & 75.0 & 78.9 \\
    \bottomrule
  \end{tabular}
\end{table}

\begin{table}[htpb]
  \centering
  \footnotesize
  \setlength{\tabcolsep}{4.5pt}
  \renewcommand{\arraystretch}{1.12}
  \caption{Per-concept \textbf{consistency assessment}. Readers judged whether the LLM-based consistency assessment between the classification label and the top-1 retrieved text was correct for each case. $N$ is the number of rated cases; Consistent, Inconsistent, and No feedback (missing or not possible) are percentages of $N$ (row-wise sum $=100\%$). Metrics are pooled across all readers.}
  \label{tab:appendix_vlm_consistency}
  \begin{tabular}{p{0.30\linewidth} rrrr rrrr}
    \toprule
    & \multicolumn{4}{c}{CheXpert} & \multicolumn{4}{c}{Internal} \\
    \cmidrule(lr){2-5}\cmidrule(lr){6-9}
    Concept & $N$ & Cons. & Inc. & No fb. & $N$ & Cons. & Inc. & No fb. \\
    \midrule
    \textbf{Support devices} & & & & & & & & \\
    \quad Airway & 100 & 96.0 & 3.0 & 1.0 & 96 & 94.8 & 3.1 & 2.1 \\
    \quad Feeding tube & 100 & 90.0 & 8.0 & 2.0 & 95 & 86.3 & 12.6 & 1.1 \\
    \quad Central venous access & 100 & 98.0 & 2.0 & 0.0 & 95 & 100.0 & 0.0 & 0.0 \\
    \quad Chest drain & 98 & 89.8 & 8.2 & 2.0 & 94 & 96.8 & 1.1 & 2.1 \\
    \quad Pacemaker & 99 & 100.0 & 0.0 & 0.0 & 95 & 95.8 & 4.2 & 0.0 \\
    \quad Other & 98 & 89.8 & 10.2 & 0.0 & 95 & 85.3 & 13.7 & 1.1 \\
    \textbf{Thoracic organs} & & & & & & & & \\
    \quad Heart & 98 & 89.8 & 5.1 & 5.1 & 94 & 88.3 & 7.4 & 4.3 \\
    \quad Mediastinum & 97 & 90.7 & 8.2 & 1.0 & 94 & 84.0 & 12.8 & 3.2 \\
    \textbf{Pathologies} & & & & & & & & \\
    \quad Lung: Pneumonia & 98 & 87.8 & 11.2 & 1.0 & 93 & 92.5 & 3.2 & 4.3 \\
    \quad Lung: Atelectasis & 98 & 93.9 & 5.1 & 1.0 & 93 & 91.4 & 6.5 & 2.1 \\
    \quad Lung: Emphysema & 97 & 82.5 & 16.5 & 1.0 & 93 & 83.9 & 9.7 & 6.5 \\
    \quad Lung: Fibrosis & 97 & 87.6 & 10.3 & 2.1 & 90 & 74.4 & 15.6 & 10.0 \\
    \quad Lung: Mass lesion & 98 & 83.7 & 13.3 & 3.1 & 92 & 82.6 & 12.0 & 5.4 \\
    \quad Pleura: Pneumothorax & 97 & 82.5 & 15.5 & 2.1 & 92 & 91.3 & 5.4 & 3.3 \\
    \quad Pleura: Pleural effusion & 98 & 94.9 & 5.1 & 0.0 & 93 & 87.1 & 9.7 & 3.2 \\
    \quad Vessels: Congestion & 98 & 95.9 & 4.1 & 0.0 & 93 & 92.5 & 6.5 & 1.1 \\
    \quad Vessels: Pulmonary edema & 98 & 88.8 & 9.2 & 2.0 & 93 & 95.7 & 3.2 & 1.1 \\
    \quad Bones: Fracture & 98 & 86.7 & 12.2 & 1.0 & 93 & 81.7 & 15.0 & 3.2 \\
    \quad Bones: Other & 95 & 85.3 & 10.5 & 4.2 & 92 & 75.0 & 19.6 & 5.4 \\
    \midrule
    \textit{Macro average} & 1862 & 90.2 & 8.3 & 1.5 & 1775 & 88.4 & 8.5 & 3.1 \\
    \bottomrule
  \end{tabular}
\end{table}
\fi

\end{document}